\documentclass{article}
\usepackage{enumitem}
\setlist{nosep, leftmargin=14pt}

\usepackage{graphicx}
\usepackage[dvipsnames]{xcolor}
\usepackage{soul}
\usepackage{amsmath}
\usepackage{siunitx}
\usepackage{amsfonts}
\usepackage{url}
\usepackage[dvipsnames]{xcolor}
\usepackage{appendix}
\usepackage[preprint]{spconf}
\toappear{\textit{Preprint submitted to NeuroImage}}

\def\x{{\mathbf x}}

\title{Motion Correction and Volumetric Reconstruction for Fetal Functional Magnetic Resonance Imaging Data}

\name{\begin{tabular}{c}Daniel Sobotka$^{1}$, Michael Ebner$^{3}$, Ernst Schwartz$^{1}$, Karl-Heinz Nenning$^{1,4}$, Athena Taymourtash$^{1}$, \\ Tom Vercauteren$^{3}$, Sebastien Ourselin$^{3}$, Gregor Kasprian$^{2}$, Daniela Prayer$^{2}$, Georg Langs$^{1}$, \\ 
and Roxane Licandro$^{1,5}$\end{tabular}}

\address{$^{1}$ Computational Imaging Research Lab , Department of Biomedical Imaging and  Image-guided Therapy, \\ Medical University of Vienna, Vienna, Austria \\
$^{2}$ Division of Neuroradiology and Musculoskeletal Radiology, Department of Biomedical Imaging \\ and Image-guided Therapy, Medical University of Vienna, Vienna, Austria \\
$^{3}$ School of Biomedical Engineering \& Imaging Sciences, King's College London, London, United Kingdom \\
$^{4}$ Center for Biomedical Imaging and Neuromodulation, Nathan Kline Institute, Orangeburg, NY, USA \\
$^{5}$ Laboratory for Computational Neuroimaging, Athinoula A. Martinos Center for Biomedical Imaging, \\Massachusetts General Hospital and Harvard Medical School, Charlestown, MA, USA }

\begin{document}
\maketitle
\begin{abstract}
Motion correction is an essential preprocessing step in functional Magnetic Resonance Imaging (fMRI) of the fetal brain with the aim to remove artifacts caused by fetal movement and maternal breathing and consequently to suppress erroneous signal correlations. Current motion correction approaches for fetal fMRI choose a single 3D volume from a specific acquisition timepoint with least motion artefacts as reference volume, and perform interpolation for the reconstruction of the motion corrected time series. The results can suffer, if no low-motion frame is available, and if reconstruction does not exploit any assumptions about the continuity of the fMRI signal. Here, we propose a novel framework, which estimates a high-resolution reference volume by using outlier-robust motion correction, and by utilizing Huber L2 regularization for intra-stack volumetric reconstruction of the motion-corrected fetal brain fMRI. We performed an extensive parameter study to investigate the effectiveness of motion estimation and present in this work benchmark metrics to quantify the effect of motion correction and regularised volumetric reconstruction approaches on functional connectivity computations. We demonstrate the proposed framework's ability to improve functional connectivity estimates, reproducibility and signal interpretability, \textcolor{Black}{which is clinically highly desirable for the establishment of prognostic noninvasive imaging biomarkers.} The motion correction and volumetric reconstruction framework is made available as an open-source package of NiftyMIC.
\end{abstract}
\section{Introduction}\label{sec:Introduction}
During gestation the fetal brain undergoes various developmental changes, guided by environmental factors and genetic programs \cite{Power2010}. These processes start in the third Gestational Week (GW) by the differentiation of neural progenitor cells and proceed in building structural as well as functional trajectories through late adolescence \cite{stiles2010basics}. Important anatomical and functional neural networks in the fetal brain are established during the second trimester of pregnancy \cite{kasprian2008utero}. 

From the $18^{th}$ GW the fetal brain maturation can be studied by in-vivo Magnetic Reso\-nance Imaging (MRI) \cite{prayer2006mri}. This image acquisition technique is considered as a safe diagnostic imaging procedure, without documented short term \cite{Gowland2010,ray2016association} or long term adverse effects \cite{zvi2020fetal}. Imaging a fetus is challenging, due to its constantly changing position and movements, which consequently can cause blurring and imaging artefacts. Thus, a main focus in creating suitable imaging protocols lies in shortening the acquisition time to reduce the impact of motion while imaging. The development of functional Magnetic Resonance Imaging fMRI) protocols for brain analysis started in the 90's, where \cite{biswal1995functional} discovered that during resting of a patient low frequency fluctuations in Blood Oxygenation Level-Depen\-dent (BOLD) signals are highly temporally correlated with motor cortex regions. A common approach to counteract motion artifacts in fMRI is to use an interleaved Echo Planar Imaging (EPI) \cite{butts1994interleaved} setup to minimize slice excitation leakage to adjacent slices of one timepoint \cite{parker2014retrospective}. There, the interleaved parameter defines how many adjacent slices of a brain are acquired at different times within a certain repetition time. A drawback of such an approach is the introduction of spatial distortions due to magnetic field inho\-mo\-ge\-neities and differences in timing across different image slices
\cite{poldrack2011handbook}. Today, the fMRI technique forms an important tool for exploring Functional Connectivity (FC) networks and its development in the human brain. In the context of motion, however, subsequent and reliable analysis of such data is challenging.

\subsection{Fetal Functional Connectivity Analysis}
While functional connectivity networks are well investigated in the adult's  \cite{sepulcre2010organization,thomas2011organization,ferreira2013resting,geerligs2014brain} and child's brain \cite{supekar2009development,fair2009functional,supekar2010development,licandro2016changing}, recently new research for the fetal brain \cite{schopf2012watching,jakab2014fetal,thomason2017weak,van2018hubs,thomason2019prenatal,wheelock2019sex,turk2019functional}, preterm brain \cite{stoecklein2020variable,de2020functional} and neonatal brain \cite{ciarrusta2020emerging} evolved. 
In fetal functional connectivity analysis, preprocessing is an essential step to remove artefacts and to suppress erroneous signal correlations caused by motion \cite{esteban2019fmriprep}. For a correct data interpretation, it is particularly important that preprocessing procedures assist to obtain an improved signal quality on the one hand, but without introducing bias on the other hand. In the adult's or child's brain it has been shown that altered functional connectivity networks can indicate mental health disorders, such as autism spectrum disorder \cite{weng2010alterations}, schizophrenia \cite{bluhm2007spontaneous} or depression \cite{greicius2007resting}. Fetal functional connectivity analysis aims to discover the functional connectome organization of the fetal brain and to identify evolving cognitive functions before birth. However, reproducible measurements for reliable connectivity analysis are especially difficult to achieve in the context of motion. Evaluation metrics can therefore be crucial to ensure sufficient signal quality is available for further analysis \cite{strother2004optimizing}. 
While preprocessing standards have been established for the adult brain \cite{esteban2019fmriprep}, motion correction of fetal fMRI and subsequent interpretation of processed data is still challenging. Recent research on the topic of fetal fMRI preprocessing introduced open-source pipelines (brain segmentation and MCFLIRT motion correction) \cite{rutherford2019observing}, but different approaches to cope with motion are established. To account for motion present during image acquisition two steps are required: First, the spatial alignment of volumes (motion correction) and second, the minimization of a nuisance term (nuisance or motion regression). Additionally, motion scrubbing (censoring) can be done to exclude high-motion corrupted time points. All the presented methods will be described in detail in the following:
\begin{figure*}
\centering
\includegraphics[width=0.95\textwidth]{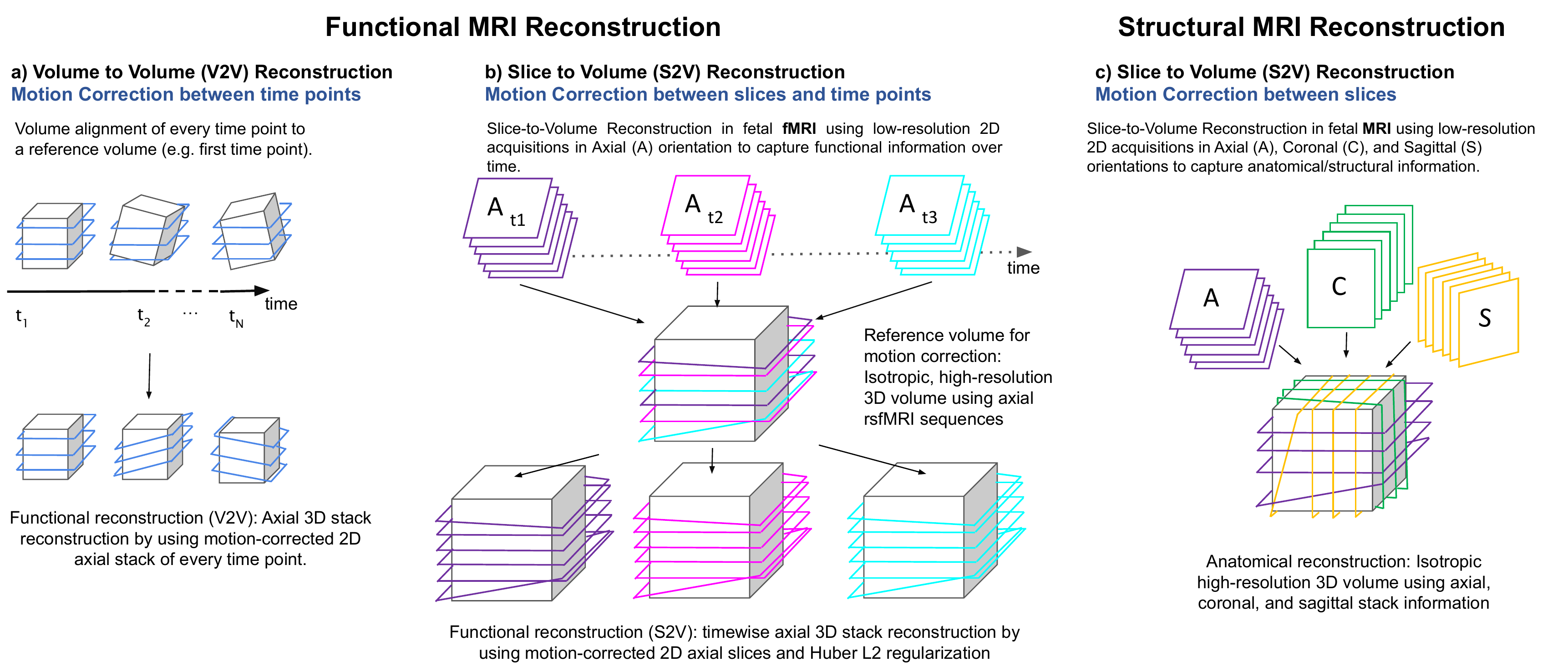}
\caption{Overview of Volume-to-Volume (V2V) and Slice-to-Volume (S2V) reconstruction in functional MRI, which uses only axial slices for reference volume computation compared to structural MRI reconstruction: (a) Existing approaches use a V2V reconstruction, where each fMRI time point (stack) is rigidly registered to a reference volume (e.g. first time point or a high-resolution representation such as proposed in this work); (b) Our proposed S2V approach with Huber L2 volumetric reconstruction: First, a high-resolution reference volume is estimated and each slice of each time point is rigidly registered individually to it. Afterwards the input slices are reconstructed with Huber L2 regularization \textcolor{Black}{; (c) In contrast to fMRI reconstruction, where only single-orientation acquisitions are available, for structural MRI, acquisitions in axial, coronal and sagittal orientation are used for reconstruction.}}
\label{fig:overview_fmri}
\end{figure*}
\\\textit{Motion correction} addresses the difficult task of removing artifacts caused by gross maternal motion, fetal movement or intrinsic movement of maternal organs and bloodflow. This movement can be described with regards to the image acquisition as either in-plane or out-of-plane. For in-plane motion, the axial head slice is rotating from left to right, while for out-of-plane motion also up and down movement of the axial head slice is visible. In the case of interleaved acquisition, these patterns can be present jointly at a single timepoint (tp) depending on the interleaved parameter. Compared to MRI motion correction, only the axial image acquisitions are used for fMRI.\\
\textit{Motion regression} or \textit{nuisance regression} is a linear regression method, where voxel-wise fMRI time series are re\-gressed against prior estimated rotational and translational head motion parameters \cite{parkes2018evaluation}. \textit{ICA-AROMA} uses independent component analysis to identify motion components \cite{pruim2015ica}, where \textit{slice timing correction} is a preprocessing step to correct for slice-dependent delays \cite{parker2019benefit}. \\
In recent fetal studies \cite{thomason2017weak,van2018hubs,thomason2019prenatal,wheelock2019sex}, \textit{motion censoring or scrubbing} instead of motion correction is used where datasets with high motion are excluded, i.e. heavily motion corrupted volumes of a 4D functional acquisition are removed. While this method works well for low motion fetal acquisitions, high motion corrupted data leads to a high degree of data exclusion and thus data loss. Motion censoring not only decreases the number of investigable time points of subjects with more movement, it can also lead to bias effects, such as correlation estimates, which become excessively noisy \cite{power2015recent}. This might be particularly problematic in settings, where clinical diagnostic imaging and resting state research protocols are combined and image acquisition times are limited and precious. \\\\
Motion correction algorithms (see Figure \ref{fig:overview_fmri}a) are available in fMRI analysis packages, such as FSL \cite{jenkinson2002improved}, AFNI \cite{cox1996afni}, and SPM \cite{friston1995spatial}. However, being developed for adult brains, they are not designed to cope with fetal movement between slice acquisitions \cite{seshamani2016detecting}. Therefore, new motion correction strategies are required to cope with motion artifacts associated with high motion and interleaved acquisition protocols typical for fetal fMRI. There has been recent research on the topic of motion correction in fMRI \cite{pinsard2018integrated,huang2020improved} and especially for fetal fMRI \cite{ferrazzi2014resting,you2016robust,scheinost2018fetal}. \cite{ferrazzi2014resting} creates first a reference image by averaging the entire dataset, where the data is divided into temporally contiguous blocks with fewer slices. Then rigid registration is performed and the reference image is subsequently updated. This process is repeated until each block contains one slice and maps all the slices with a rigid transformation from the stationary scanner coordinate system into the moving fetal brain reference system. Compared to our method the volumetric reconstruction step is not based on outlier-robust super resolution reconstruction. \textcolor{Black}{\cite{you2016robust} differentiate between local and global motion in their approach. First, the timeseries is splitted into smaller blocks and registration is performed independently within each block. Second, the field of view for image registration is restricted to the region of interest. Afterwards, volume outlier rejection, voxel outlier rejection and imputation of missing data is done. \cite{scheinost2018fetal} extracts $2\textsuperscript{nd}$ order edge features from the fetal brain to perform a weighted rigid registration with normalized correlation to align the edge features of each timepoint to a reference image.} \cite{pinsard2018integrated} uses an extended Kalman Filter for estimation of slice-wise motion, where \cite{huang2020improved} uses boundary-based registration for motion correction of adult fMRI. 
\subsection{Contribution}
Motion-robust super resolution reconstruction models developed for structural fetal MRI (see Figure \ref{fig:overview_fmri}c) are not applicable for fMRI due to slices acquired only in a single axial direction, rather than multiple directions typically including multiple axial, coronal, and sagittal plane acquisitions. In addition, image resolution in fMRI is worse in both in- and through-plane direction compared to structural MRI, since a higher temporal resolution is required to measure brain activity over time. In Volume-to-Volume approaches (V2V) each time point is aligned to a reference, where in Slice-to-Volume (S2V) approaches each slice of a time point is aligned individually.
Available motion correction algorithms from fMRI analysis packages (see Figure \ref{fig:overview_fmri}a) use a V2V approach, where each time point is registered rigidly to a target time point, and cannot cope with interleaved slice acquisitions. Current methods are restricted to choose an existing timepoint characterised by low motion, which is not ideal or might not exist. We hypothesise that improved motion correction per time point can be achieved if a high-resolution (HR) reference volume is available. Thus, in this work we propose a framework (see Figure \ref{fig:overview_fmri}b), which estimates a high-resolution reference using outlier-robust motion correction and super-resolution reconstruction in a first step. Subsequently, the computed reference volume is used to motion-correct each individual time point. Alternatively, other methods such as \cite{huang2020improved} use structural MRI data as reference for motion correction. A disadvantage of this strategy lies in a different MRI image contrast, which aggravates the registration problem to a multi-modal one. By introducing a high-resolution reference, a mono-modal registration of fMRI images of the same contrast can be performed.

To counteract fetal movement, the fMRI dataset used in this work is acquired with an interleaved EPI, where each 3D stack $\mathbf{t}_x$ is composed of slices acquired at three different time points $\mathbf{t}_{x_a}$, $\mathbf{t}_{x_b}$ and $\mathbf{t}_{x_c}$. For the evaluation of the effectiveness of the proposed motion correction technique, we created synthetic datasets to simulate interleaved rotation and translation motion in $x$, $y$ and $z$ direction. Furthermore, we performed a parameter study to estimate the motion regularization parameters for Huber L2, first-order Tikhonov L2 , total variation L2 regularization after motion correction. Since the S2V motion correction from \cite{ferrazzi2014resting} is not publicly available, we used the MCFLIRT \cite{jenkinson2002improved} motion correction technique\textcolor{Black}{, which is also used in the fetal fMRI pipeline from \cite{rutherford2019observing},} as a baseline to compare the results from the motion correction and volumetric reconstruction framework proposed. We used reproducibility measures to quantify the effect of motion correction and regularization on the fMRI signal. Furthermore, intensity-based measurements are used to assess signal similarity between timepoints and reproducibility of functional connectivity between two time courses of each subject. \\
The contribution can be summarized as follows:
\begin{itemize}
    \item Outlier-robust super resolution approach for high-res\-o\-lu\-tion reference volume estimation for slice-based fMRI motion correction.
    \item Huber L2 regularization term for the volumetric reconstruction process.
    \item Evaluation scheme to assess the performance of the motion correction approach proposed by using synthetic test data and an extensive parameter study.
    \item Signal quality assessment strategy to determine the impact of the proposed regularisation approach associated with our volumetric reconstruction strategy for motion-corrected fMRI data on functional connectivity computations.
    \item Comparison of the proposed motion correction and volumetric reconstruction framework with available MCFLIRT motion correction from the FSL\footnote{\url{https://fsl.fmrib.ox.ac.uk/fsl/fslwiki/MCFLIRT} [accessed 2022-2-2]} fMRI package.
\end{itemize}
Results show that with the motion correction and volumetric reconstruction approach fewer timepoints of datasets are detected as outliers, standard deviation of brain voxels are reduced and comparability and reproducibility of functional connectivity estimates are increased compared to the uncorrected input and MCFLIRT motion correction. This improves the efficiency of acquired fetal fMRI data and allows to optimally use them for diagnostic purposes in a clinical research setting.

\section{Materials and Methods}\label{sec:Methodology}
\def\x{\mathbf{x}}
\def\y{\mathbf{y}}
\def\AA{\mathbf{A}}
\def\nnabla{\boldsymbol{\nabla}}
\newcommand{\norm}[2][]{#1\Vert#2#1\Vert}
In this section, the proposed motion correction and volumetric reconstruction framework is introduced. Subsequently, the used evaluation metrics are summarized as well as the functional connectivity computation and the assessment of its reproducibility, is presented. % 
\\
\\
\textbf{Data.}
The study includes a total of $15$ \textcolor{Black}{axial} fMRI sequences from fetuses between the 20\textsuperscript{th} and 37\textsuperscript{th} GW (mean: $27.93$, standard deviation: $5.2$) with normal brain development. \textcolor{Black}{All the subjects are control cases from an in-house clinical study and there was no motion-related exclusion criteria.}
Functional MRI was performed on a \SI{1.5}{T} clinical scanner (Philips Medical Systems, Best, The Netherlands) using a SENSitivity Encoding (SENSE) cardiac coil with five elements (three posterior, two anterior) wrapped around the mother's abdomen, utilizing single-shot gradient-recalled echo-planar imaging and no cardiac gating with the following setup: \SI{50}{\ms} echo time, \SI{1000}{\ms} repetition time, \SI{1.736}{\mm} in-plane resolution, \SI{3}{\mm} through-plane resolution, \SI{18} slices for each of the \SI{96} vol\-umes. The pregnant women were examined in the supine or left decubitus position (feet first), and no contrast agents or sedatives were administered. In order to receive the optimal MR signal, the coil was readjusted depending on the position of the fetal head during the imaging procedure. The local ethics committee approved the protocol of this study, which was performed in accordance with the Declaration of Helsinki.
\textcolor{Black}{Additionally, the proposed framework has been evaluated with an in-house 3T dataset to evaluate the performance and generalizability to other protocols and scanner types. The data included $6$ axial sequences from two different scanners (Philips Ingenia Elition X and Philips Achieva) with the following setup: \SI{50}{\ms} echo time, \SI{1000}{\ms} repetition time, \SI{1.667}{\mm} in-plane resolution, \SIrange{3}{5}{\mm} through-plane resolution, \SI{17} to \SI{20} slices for each of the \SI{96} vol\-umes.} \textcolor{Black}{Brain masking was performed with the Brain Extraction Tool \cite{smith2002fast} incorporated in the FSL toolbox \footnote{\url{https://fsl.fmrib.ox.ac.uk/fsl/fslwiki/BET} [accessed 2022-2-2]}. Especially in low gestational age subjects the brain masking fails sometimes to differentiate between fetus and placenta, therefore manual fine tuning of all brain masks are accomplished. The input of the framework are two four dimensional arrays containing (1) a sequence of 3D fMRI stacks acquired over time, where each stack consists of low-resolution 2D slices and (2) a corresponding fetal brain masks.}
\subsection{Motion correction and volumetric reconstruction framework} \label{fsrr-section}
\begin{figure}%[!h]
\centering
\includegraphics[width=0.8\columnwidth]{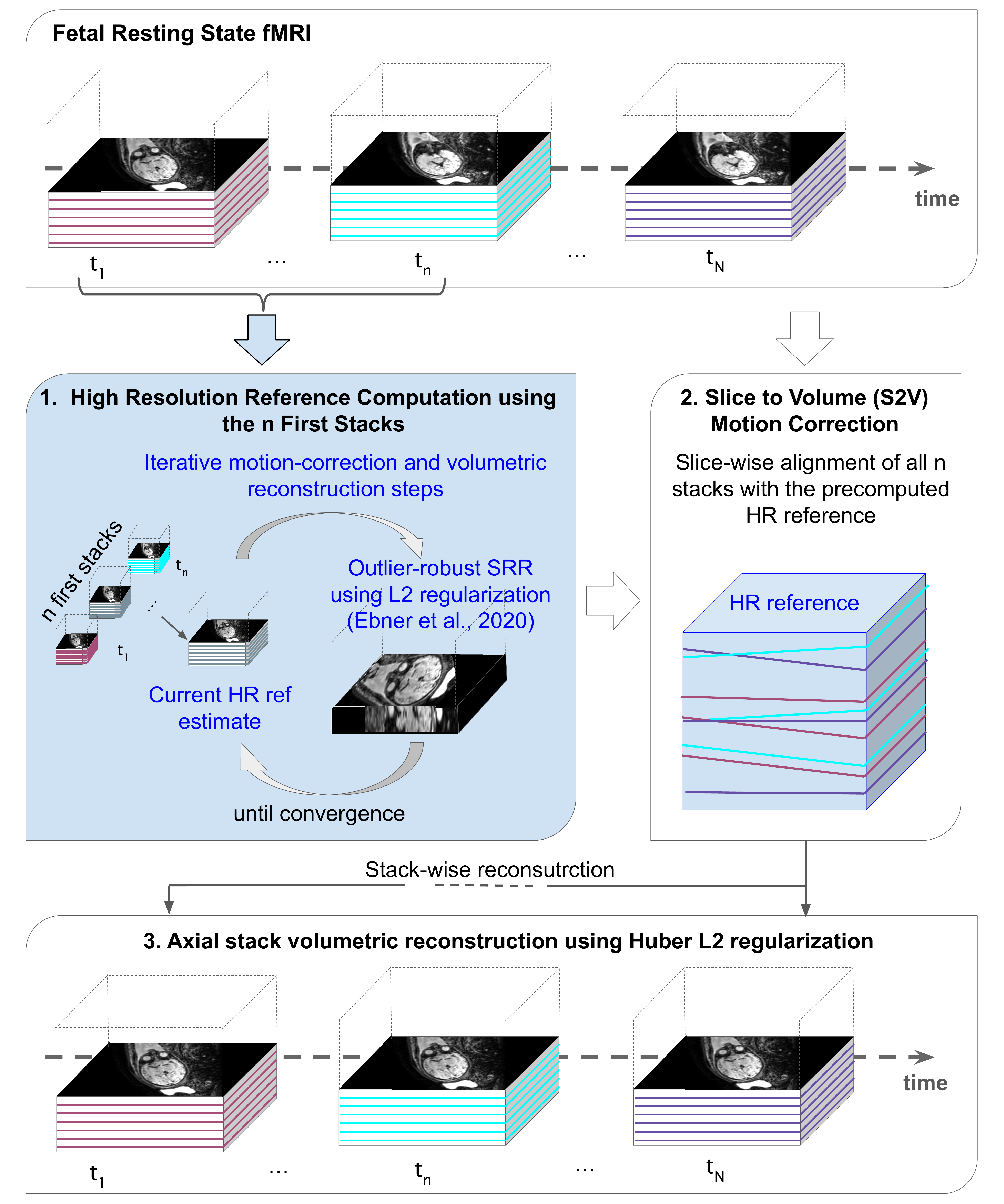}
\caption{Overview of the proposed motion correction and volumetric reconstruction algorithm for fMRI. With the first $x$ time points a HR reference volume with outlier rejection is estimated \cite{ebner2020automated}. Afterwards each slice of a time point is registered to the HR reference volume and with Huber L2 regularization the input slices are reconstructed.}
\label{fig:algorithm}
\end{figure}

The framework introduced consists of three parts, which are also visualized in Figure \ref{fig:algorithm}: (1) Outlier-robust high-reso\-lution reference volume construction, (2) Rigid slice-based motion correction and (3) Huber L2-based volumetric reconstruction. 
The whole algorithm is available as part of the open-source reconstruction framework NiftyMIC\footnote{available in release version 0.9 at \url{https://github.com/gift-surg/NiftyMIC}} and we will describe the components in detail in the following sections.
\subsubsection{Outlier-robust high-resolution Reference Volume Construction}
The first step of the framework proposed creates a high-resolution reference 3D volume from the first $n$ stacks of an fMRI fetal acquisition using outlier-robust motion-correc\-tion and volumetric reconstruction based on L2-regularisation \\ \cite{ebner2020automated}.
\subsubsection{Rigid slice-based motion correction}\label{sec:moco_sec}
In the rigid slice-based motion correction step V2V is performed to estimated HR reference volume using symmetric block matching \cite{modat2014global}. Afterwards S2V is performed to the estimated HR reference volume with normalised cross-correlation to rigidly register each slice of a time point individually to the high-resolution reference volume.%
\subsubsection{Huber L2-based Volumetric reconstruction}
The volumetric reconstruction for axial volume reconstruction per time point using regularisation is expressed in Eq.~\ref{eq:optimization}, where $K$ denotes the total number of slices per time point, $A_k$ denotes the linear operator, $\vec{\xi}$ the reconstructed volume, $\vec{y_k}$ the input slice after motion correction described in section \ref{sec:moco_sec} and $\mathcal{R}$ the regularization term. For each time point, a subject volumetric reconstruction is performed separately. Subsequently, the volume at the original axial grid per time point using Huber L2 (Eq.~\ref{eq:huberl2}) regularization, where $\alpha\geq 0$ denotes the regularization parameter, $\nnabla$ the differential operator and $\gamma$ the parameter for the Huber norm. After each time point is reconstructed individually, a motion-corrected 4D fMRI volume is obtained (for further analysis).
\begin{equation}\label{eq:optimization}
    \vec{x} := \mathrm{argmin}_{\vec{\xi}\in\mathcal{R}^{N}} \Big[ \sum_{k=1}^{K} \norm{A_k\vec{\xi} - \vec{y_k}}_{\ell^2}^2 + \mathcal{R(\vec{\xi})} \Big]
\end{equation}
\begin{equation}\label{eq:huberl2}
    \mathcal{R}_{Hu}(\vec{\xi}) = \frac{1}{2\alpha} \norm{\nnabla\vec{\xi}}_{\gamma} = \begin{cases}
\nnabla\vec{\xi} & \nnabla\vec{\xi} < \gamma^2 \\ \\
2\gamma\sqrt{\nnabla\vec{\xi}}-\gamma^2 & \nnabla\vec{\xi} \geq \gamma^2
\end{cases}
\end{equation}
\subsection{Metrics to assess motion correction}
In this work, a total of four metrics is used to assess the quality of the obtained reconstructions after motion correction and volumetric reconstruction and to measure the reproducibility of functional connectivity computations between two time courses of a subject. Motivated by the fact that motion and artifacts in fMRI acquisitions lead to signal intensity changes, \textit{standard deviation} of the signal over time is used as first metric for the evaluation of the motion correction and volumetric reconstruction framework. As second metric the \textit{Structural Similarity Index Measure (SSIM)} \cite{wang2004image} is computed to assess the difference between the timepoints of a fMRI sequence, where a value up to $1$ indicates similarity and $0$ no similarity. Both metrics have drawbacks, when smoothing is applied or an artifact has similar signal intensity compared to brain activity patterns. Thus, as third metric, we use \textit{outlier ratio} to measure the percentage of outlier voxels for each subject's acquisition timepoint after motion correction and volumetric reconstruction. For the development of the fourth metric we hypothesize the following: Given a motion corrupted fMRI time course and dividing it in two halves, the connectivity computations performed on these two half time courses become more similar if the removal of motion was successful. %
Thus, as fourth metric \textit{functional connectivity reproducibility} measurements are used to evaluate the effect from the motion correction and volumetric reconstruction on cortical connectivity patterns on the fetal brain surface. 
\subsubsection{Standard Deviation Measurements} \label{std-metric}
For every brain voxel $y$ in an image $I$ the standard deviation value over time is computed. Increased standard deviation values ($> 0$) indicate lower similarity, since we hypothesize that pixel variations will be reduced in a successful case of motion correction and reconstruction. Pixel variations may occur due to either reconstruction or motion correction insufficiencies, but also due to inherent variability in the signal. Therefore, this metric serves only for basic analysis. 

\subsubsection{Outlier ratio}
Outlier voxels $y_{o}$ of a time series are defined as follows \cite{cox1996afni}:
\begin{equation}\label{eq:outlier_voxel}
y_{o} = \textrm{qinv}\left(\frac{0.001}{N}\right) \cdot \sqrt{\frac{\pi}{2}} \cdot \textrm{MAD}
\end{equation}
\textrm{MAD} denotes the Median Absolute Deviation of the time series, \textrm{qinv} the inverse of the reversed Gaussian cumulative distribution function and $N$ the length of the time series. We define a timepoint of a subject's acquisition as rejected, if this timepoint has more than $3\%$ (similar to \cite{turk2017spatiotemporal}) outlier voxels $y_{o}$. Afterwards, the \textit{outlier ratio} is calculated by the percentage of rejected time points from the time series.
\subsubsection{Functional Connectivity}\label{sec:FC}
\begin{figure}
\centering
\includegraphics[width=0.9\columnwidth]{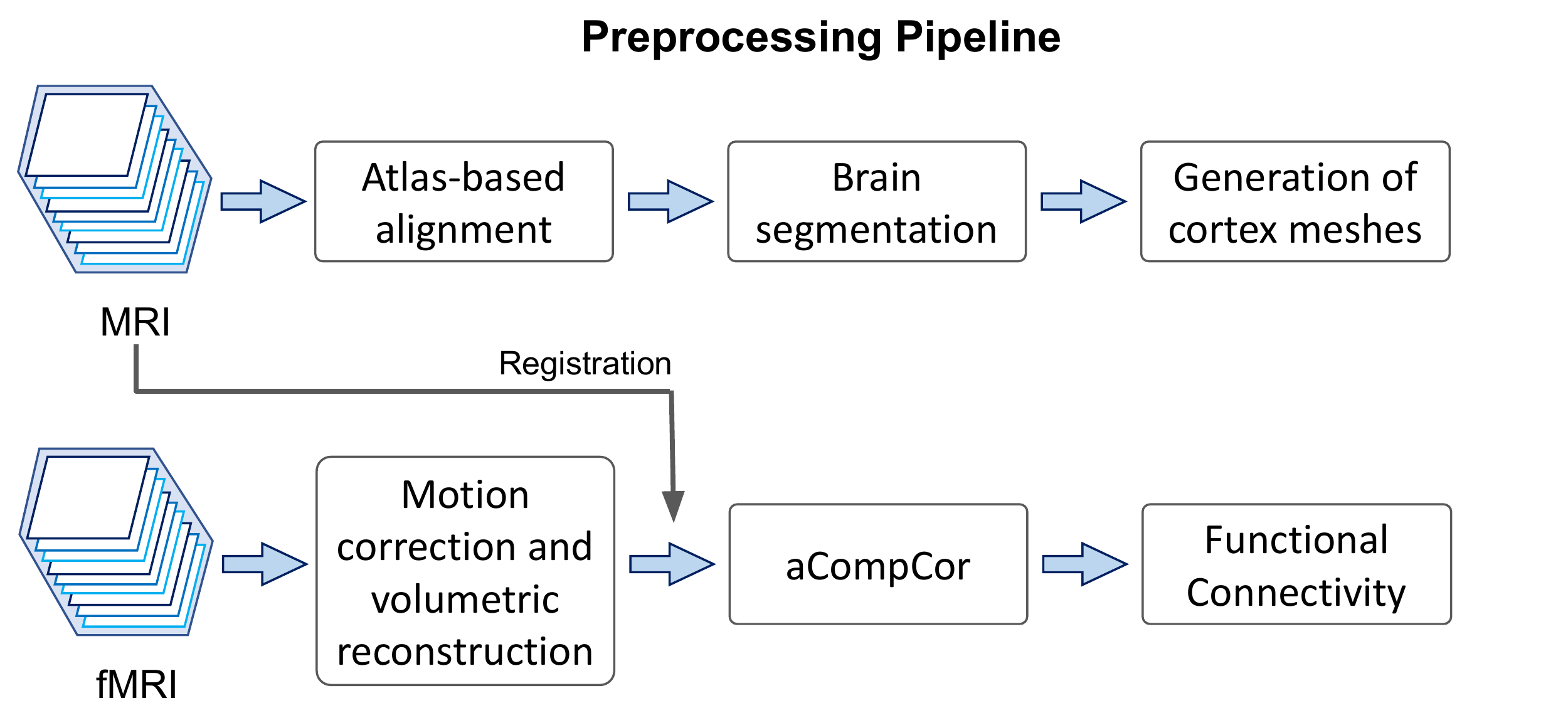}
\caption{Block diagram of the used structural MRI and fMRI preprocessing pipeline from Section. MRI acquisition's are needed to generate white matter and cerebrospinal fluid segmentations for aCompCor and cortex meshes for being able to compute the functional connectivity reproducibility metrics.}
\label{fig:pp}
\end{figure}

The preprocessing pipeline used for functional connectivity computation is illustrated in Figure \ref{fig:pp}. Neither bias field nor slice timing correction is performed to avoid altered signals and noise properties in the comparison between the raw signal and the motion corrected signal. Anatomical component based method (aCompCor, \cite{behzadi2007component}) as denoising strategy is chosen to correct motion effects on correlation values \cite{taymourtash2019quantifying}. For functional connectivity estimation the structural MRI data (axial, coronal and sagittal T2 weighted MRI was performed on a \SI{1.5}{T} clinical scanner (Philips Medical Systems, Best, The Netherlands) with the following setup: \SIrange{100}{400}{\ms} echo time, \SIrange{6000}{22000}{\ms} repetition time, \SIrange{3}{4}{\mm} slice thickness and \SI{18} to \SI{25} slices) of every subject is preprocessed by performing atlas-based alignment, brain segmentation, generation of cortex meshes \cite{schwartz2016modeling} and by registering it to the corresponding fMRI acquisition of the fetus.%
After the application of the proposed motion correction and volumetric reconstruction approach, the fMRI sequences are postprocessed with aCompCor, where the five principal components from white matter and cerebrospinal fluid signals are regressed out. \\
Subsequently, the data is resampled onto the age-dependent cortex meshes consisting of $642$ nodes and the Pearson correlation coefficient is computed between the time course $t$ ($t = 1,$\dots$,M$, where $M$ is the number of time frames) of each brain node $z_{i}(t)$ and $z_{j}(t)$ ($i,j$ = $1,\dots,N$; $N$ is the number of nodes observed)\cite{sepulcre2010organization,licandro2016changing} resulting in a correlation matrix $\mathbf{CM^S}\in\mathbb{R}^{N \times N}$ for every subject $S$:
\begin{equation}\label{eq:CM}
\mathbf{CM^S} = \frac{\sum[(z_{i}(t)-\bar{z_{i}})(z_{j}(t)-\bar{z_{j}})]}{\sqrt{\sum[(z_{i}(t)-\bar{z_{i}})^2(z_{j}(t)-\bar{z_{j}})^2]}}
\end{equation}
$\bar{z}_{i}$ denotes the mean node intensities across all time points at position $i$ and $\bar{z}_{j}$ at position $j$ respectively. \\
\\
\textbf{Correlation Difference.}
For computing the correlation difference metric first a subject's fMRI signal is divided into two parts $u$ and $v$, where $u$ denotes the first half of the time series and $v$ the second half of the time series. Subsequently, for every subject $S$ correlation matrices $\mathbf{CM^S_u}$ and $\mathbf{CM^S_v}$ are computed based on the time frames $u$ and $v$ respectively. 
For being able to assess this effect, the mean difference of correlations ($\Delta C$) between a subject's $S$ extracted correlation matrices $\mathbf{CM^S_u}$ and $\mathbf{CM^S_v}$ is computed. \\
\\
\textbf{Degree Value.}
The degree value metric aims to determine the impact of motion correction and regularization values on functional connectivity estimates, which are sensitive to head motion \cite{pujol2014does}. For every cortical node $i$, high correlating time courses (threshold $\geq$ 0.3, similar to \cite{licandro2016changing}) are assigned to the degree value:
\begin{equation}\label{eq:degree}
D_i = \sum d_{ij},  i,j=1\dots N, i \neq j,
\end{equation}
The determination of $D_i$ reveals how many cortical nodes are functionally connected with this node $i$.
\section{Experimental Design and Results}\label{sec:Results}
The experimental setup of this work focuses on three aspects: We first evaluate the performance of the proposed motion correction technique for fetal fMRI. Second, we perform a parameter study for proposed regularisation strategy for the fMRI volumetric reconstruction step, to assess the effect of different regularizers on the fMRI signal. Third, we evaluate the effect of the proposed motion correction and volumetric reconstruction pipeline on fetal brain surface-based computations of functional connectivity. 
\subsection{Experiment 1: 4D Motion Correction Parameter Estimation}
In this experiment, the performance of the introduced fetal head motion estimation at the basis of motion correction is evaluated. 
\subsubsection{Experimental design}
First, a high-resolution reconstructed structural MRI volume ($1\times1\times1$ pixel spacing) from one subject of this work is created \cite{ebner2020automated}. Based on preliminary results, we used the first $15$ time points of each fMRI acquisition to calculate the HR reference volume. For the simulation of head motion patterns the following assumptions are made: The fetal head has a restricted possible range of head movement (according to the heads ability to move, womb boundaries), movement speed (speed does not exceed) and sequence (movement is continuous). To mirror these patterns in our experiments, we modelled the change of transformation (rotation and translation) parameters over time following a sinusoidal wave with different amplitudes to control the velocity of change. To imitate an interleaved movement pattern, different movement trajectories are mixed together in one time point. As example: Given a repetition time of 3 a movement parameter change for a stack is encoded as a sinusoidal wave $sin(t)$, where movement parameters $sin(1,\dots,\frac{t}{3})$ are applied to slice 1,4,7... the parameters from $sin(\frac{t}{3},\dots,\frac{2t}{3})$ to slice 2,5,8... and the parameters from $sin(\frac{2t}{3},\dots,t)$ to the remaining slices (slice 3,6,9...). A descriptive illustration is provided in Figure S1 in the supplementary material. \\ 
\begin{figure*}
\centering
\includegraphics[width=1.0\textwidth]{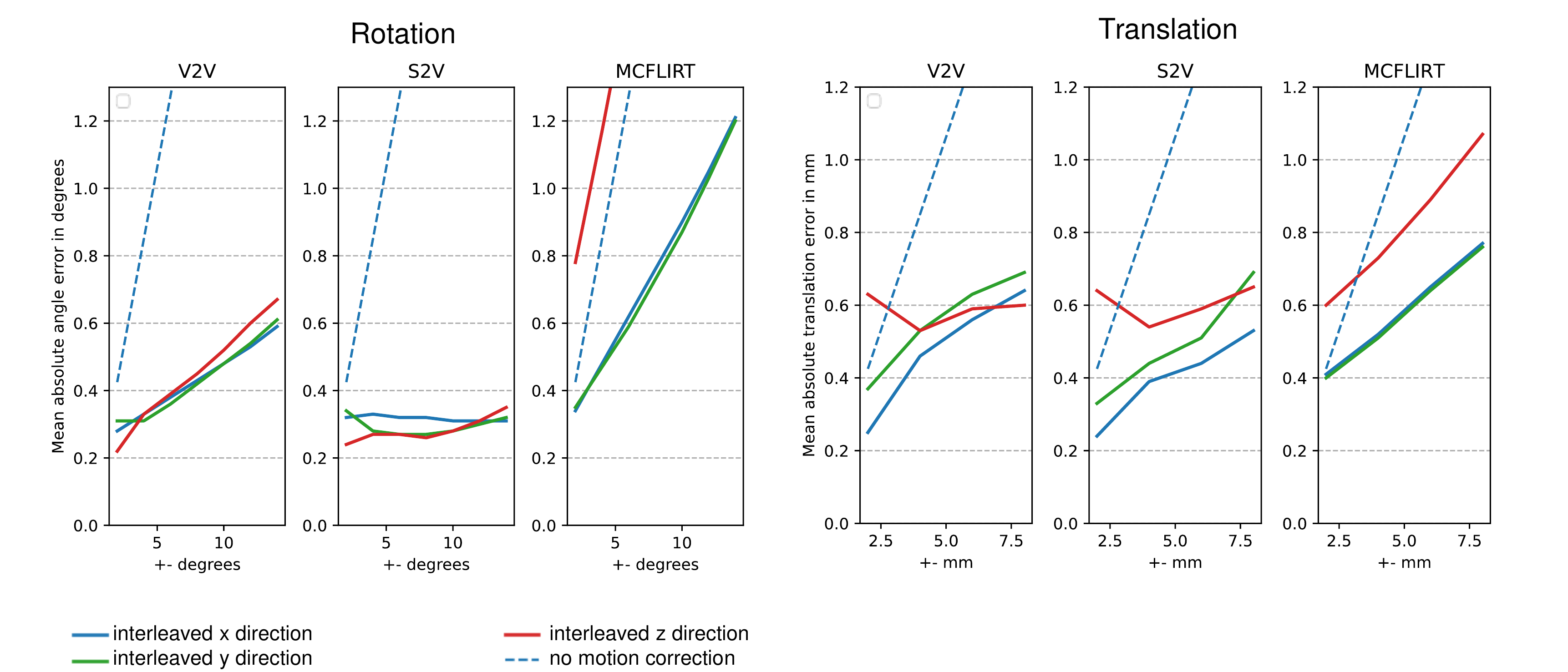}
\caption{Visualization of the Mean Absolute Error (MAE) for obtained rotation (in degrees) and translation (in mm) parameters for V2V, S2V and MCFLIRT. The blue line shows the MAE in $x$ direction, the NavyBlue line in $y$ direction and the red line in $z$ direction. The blue dashed line shows the MAE if no motion correction is performed.}
\label{fig:moco_eval}
\end{figure*}
To obtain a simulated fMRI sequence, a high-resolution reconstructed MRI volume is continuously rotated and subsequently translated for every simulation time point by applying motion parameters derived from the time-course of a sinusoidal wave. Subsequently, the obtained volumes are downsampled to a lower resolution ($1.736\times1.736\times3$ pixel spacing, \textcolor{Black}{which matches the spacing of the $1.5$T acquisitions used in this work}) sequence of volumes to simulate the lower resolution of an fMRI acquisition. With that, the motion correction estimation algorithm is executed with synthetic datasets for different rotation angles (degrees) and translation distances (mm). We used a range from $0$ to $\pm14$ degrees for rotation and from $0$ to $\pm8$ mm for translation. Higher values are excluded, since parts of the brain are moving out of the sampling grid. The synthetically created data is used to quantify the performance and effectiveness of the V2V and S2V motion correction algorithm and to simulate motion patterns in a controlled way. During evaluation the estimated rotation and translation transformation parameters are compared against the target parameters to obtain the mean absolute angle and translation difference. \textcolor{Black}{Further, to simulate $3T$ fMRI acquisitions, we conducted a second subexperiment with decreased pixel spacing ($1.536\times1.536\times3$ pixel spacing).} 
\subsubsection{Discussion and Results}
Figure \ref{fig:moco_eval} shows the mean absolute angle and translation error for V2V, S2V and MCFLIRT with in $x,y$ or $z$ direction rotated/translated synthetic interleaved data. For rotation, V2V and S2V achieves a similar mean absolute error (MAE) until $\pm4$ degrees, while with more motion the S2V approach performs better. In contrast MCFLIRT shows for every result higher error values as S2V and V2V. 
Similar results are achieved for translation, except for the $\pm14$ degrees in z direction, where V2V has a lower mean absolute translation error. 
For all synthetic rotation datasets S2V ($x$: $0.32$ $(0.01)$, $y$: $0.29$ $(0.02)$, $z$: $0.28$ $(0.03)$) showed the smallest mean and standard deviation values from MAE compared to V2V ($x$: $0.43$ $(0.10)$, $y$: $0.43$ $(0.11)$, $z$: $0.45$ $(0.14)$) and MCFLIRT ($x$: $0.77$ $(0.29)$, $y$: $0.75$ $(0.28)$, $z$: $1.99$ $(0.80)$). For the translation experiment S2V showed lower error values (S2V $x$: $0.4$ $(0.11)$, $y$: $0.49$ $(0.13)$, $z$: $0.61$ $(0.04)$) compared to V2V ($x$: $0.48$ $(0.15)$, $y$: $0.55$ $(0.12)$, $z$: $0.59$ $(0.04)$) and MCFLIRT ($x$: $0.59$ $(0.14)$, y: $0.58$ $(0.14)$, z: $0.83$ $(0.18)$). \textcolor{Black}{In the supplementary material results for the 3T pixel spacing, which are similar to the results here, are visualized (see Figure S2).}
\subsection{Experiment 2: Effect from Regularization during Reconstruction} 
In this experiment we evaluate the effect from Huber L2 and other first-order Tikhonov (TK1 L2) and total variation (TV L2) regularization for the proposed S2V motion correction by using the structural similarity index, standard deviation and outlier ratio metrics. 
\subsubsection{Experimental design}
\begin{figure}[htb]
\centering
\includegraphics[width=1\columnwidth]{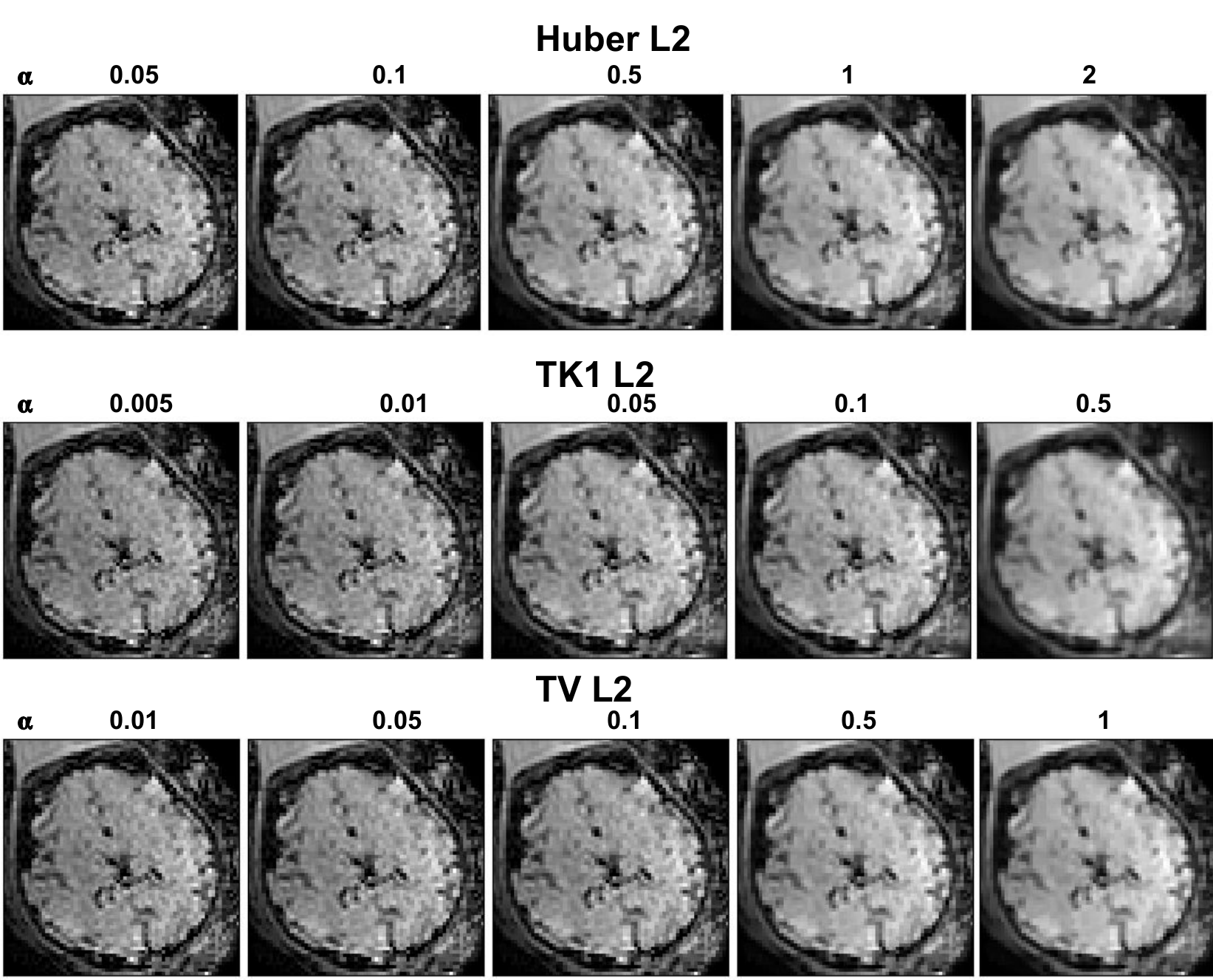}
\caption{Reconstructed images after regularization with the parameter values analyzed in this work for Huber L2, TK1 L2 and TV L2}
\label{fig:regularization}
\end{figure}

Since regularization is necessary when solving ill-posed inverse problems, we first perform \textbf{L-curve studies} \cite{hansen1999curve} to find the optimal Huber, first-order Tikhonov and total variation regularization parameter values for the reconstruction of the time points after motion correction. To be able to obtain comparability between the approaches only once the data was motion corrected in a first step and subsequently the three regularization schemes were applied separately to the same motion corrected but not reconstructed fMRI acquisition. For the L-curve studies with different regularization parameter values (from $0.001$ to $1000$), the norm of the regularized solution is plotted versus the norm of the residual norm. The estimated curve shows the optimal parameter in the edge of the curve. After L-curve studies of synthetic and real datasets we choose to analyze regularization parameter values between a range from $0.05$ and $2$ for Huber L2, from $0.005$ to $0.5$ for TK1 L2 and between $0.01$ and $1$ for TV L2. The results of the L-curve studies are added to the supplementary material (see Figure S3). %
Values higher as the chosen regularization parameter values are excluded for further analysis due to high smoothing effects on the imaging data. \\
The effect on the reconstructed images from one dataset used in this work is visible in Figure \ref{fig:regularization}.
In all cases very low parameter values (Huber L2: $0.05$, TK1 L2: $0.005$, TV L2: $0.01$) lead to sharp images, while with higher parameter values (Huber L2: $>0.5$, TK1 L2: $>0.1$, TV L2: $>0.5$) the reconstructed images' appearance are smoother. For further and better visualization for each regularization type only two parameters, which show in the outlier ratio and correlation differences the best results, are visualized.
\subsubsection{Results}
\begin{figure*}
\centering
\includegraphics[width=0.9\textwidth]{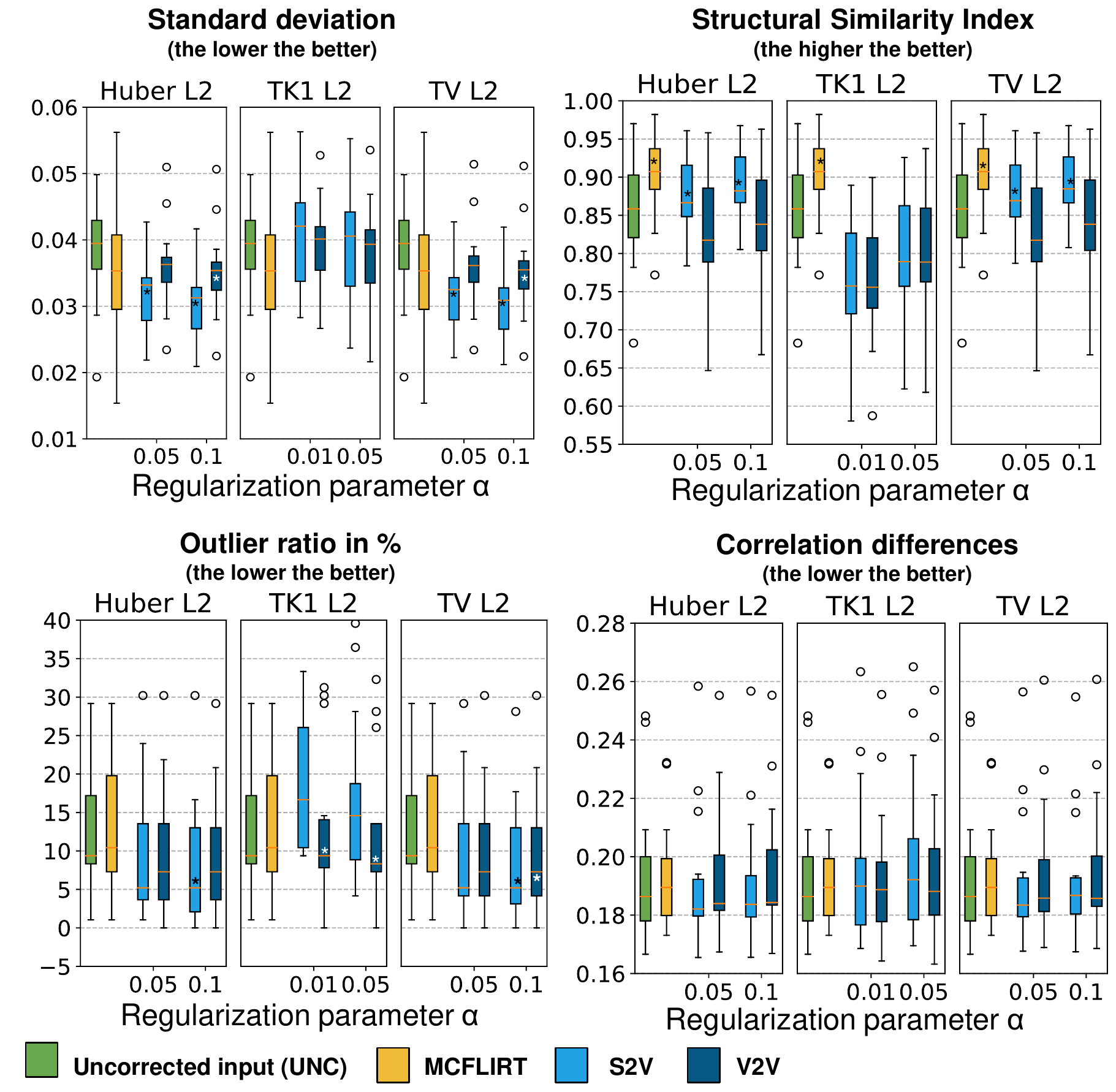}
\caption{Boxplots for the four investigated metrics over the whole population for multiple motion correction strategies (uncorrected input (UNC), MCFLIRT, V2V, and S2V) for different regularization parameters associated with TV L2, TK1 L2 and Huber L2. The top left box row shows the results from the standard deviation metric, the top right box row visualises the results from the structural similarity index metric. In the bottom left box the numbers of the rejected frames in percent are visible and in the bottom right box the correlation differences between the two investigated time ranges $u$ and $v$ are visualized. Additional evaluations are added to the supplementary material (Figure S4). Asterix symbols indicate statistical significance between UNC.}
\label{fig:boxplots}
\end{figure*}

Due to the fact that motion and motion artifacts lead to intensity changes in image voxels, \textbf{standard deviation measurement} is used to compare this effect in dependence of different regularization parameters. Figure \ref{fig:boxplots} shows in the left top box the mean standard deviation value for every subject's acquisition over the whole population.
First, every subject's acquisition is normalized between $0$ and $1$ and then the standard deviation value is calculated for every brain voxel as introduced in Section \ref{std-metric}. Afterwards the mean value over all brain voxels is visualized. Results for Huber L2 and TV L2 are showing that S2V has less standard deviation compared to V2V, MCFLIRT and the uncorrected input. In contrast for TK1 L2, V2V shows better results than S2V.\\ 
For the \textbf{SSIM metric} every fMRI timepoint $t$ of a subject's acquisition is compared with their neighboring timepoint $t+1$ and the mean value over all timepoints is visualized. Results (see top right box in Figure \ref{fig:boxplots}) demonstrate that with an increased regularization parameter a higher similarity is achieved. For Huber L2 and TV L2, S2V shows better results, where similarity is higher for every regularization parameter compared to the uncorrected input. In contrast Huber L2 V2V and both TK1 L2 results show lower similarity compared to the uncorrected input.\\ 
To investigate how many reconstructed time points are considered unreliable after the proposed motion correction and volumetric reconstruction algorithm, \textbf{outlier ratio} is used. Therefore, every fMRI timepoint of a dataset with more than three percent of voxels which are marked as outliers (Eq. \ref{eq:outlier_voxel}) are rejected. Figure \ref{fig:boxplots} shows on the bottom left box the number of rejected timepoints in $\%$ for the uncorrected input compared to TK1 L2, TV L2 and Huber L2 regularization and MCFLIRT. While for TK1 L2 S2V the number of rejected time points per dataset is higher compared to the uncorrected input (mean: $12.9\%$, median: $9.4\%$), for V2V the number of rejected volumes is lower. Huber L2 regularization (best with $0.1$ S2V, mean: $8.6\%$, median: $5.2\%$) reduces the number of rejected volumes.
Similar results are achieved with TV L2 (best with $0.05$ S2V, mean: $8.1\%$, median: $7.3\%$). \textcolor{Black}{In the supplementary material results of the standard deviation, structural similarity index and outlier ratio computation for the 3T datasets are provided (see Figure S5).} In contrast to the three metrics used, the \textbf{correlation differences} are evaluated on the fetal brain surface directly. The results from this metric are described in Section \ref{sec:Exp3}.
\subsection{Experiment 3: Surface-based Evaluation of Functional Connectivity}\label{sec:Exp3}
The focus of this experiment is to evaluate the effect from the different regularization parameter values on functional connectivity estimates on the perspective of signal reproducibility. 
\subsubsection{Experimental design}
\textbf{QC-FC (Quality Control - Functional Connectivity)} correlations \cite{power2015recent} are computed to determine how functional connectivity is modulated by a subject's movement. Therefore, we downsampled each brain's hemisphere surface representation from $642$ to $42$ nodes due high pixel spacing and computing costs.\\
We analyzed qualitatively the appearance of \textbf{correlation matrices, carpet plots and degree} value surface visualisation for UNC (row 1), MCFLIRT (row 2), Huber L2 $0.1$ S2V (row 3), TV L2 $0.05$ S2V (row 4) and TK1 L2 $0.05$ V2V (row 5) corrected signals after the application of aCompCor. With this we clean the signal from anatomical source noise signals. The correlation matrices and carpet plot calculations are performed on the entire signal time course, while the degree value surface visualisations are computed for the two time-ranges $u$ and $v$.
\subsubsection{Results}

\begin{figure*}%[!h]
\centering
\includegraphics[width=\textwidth]{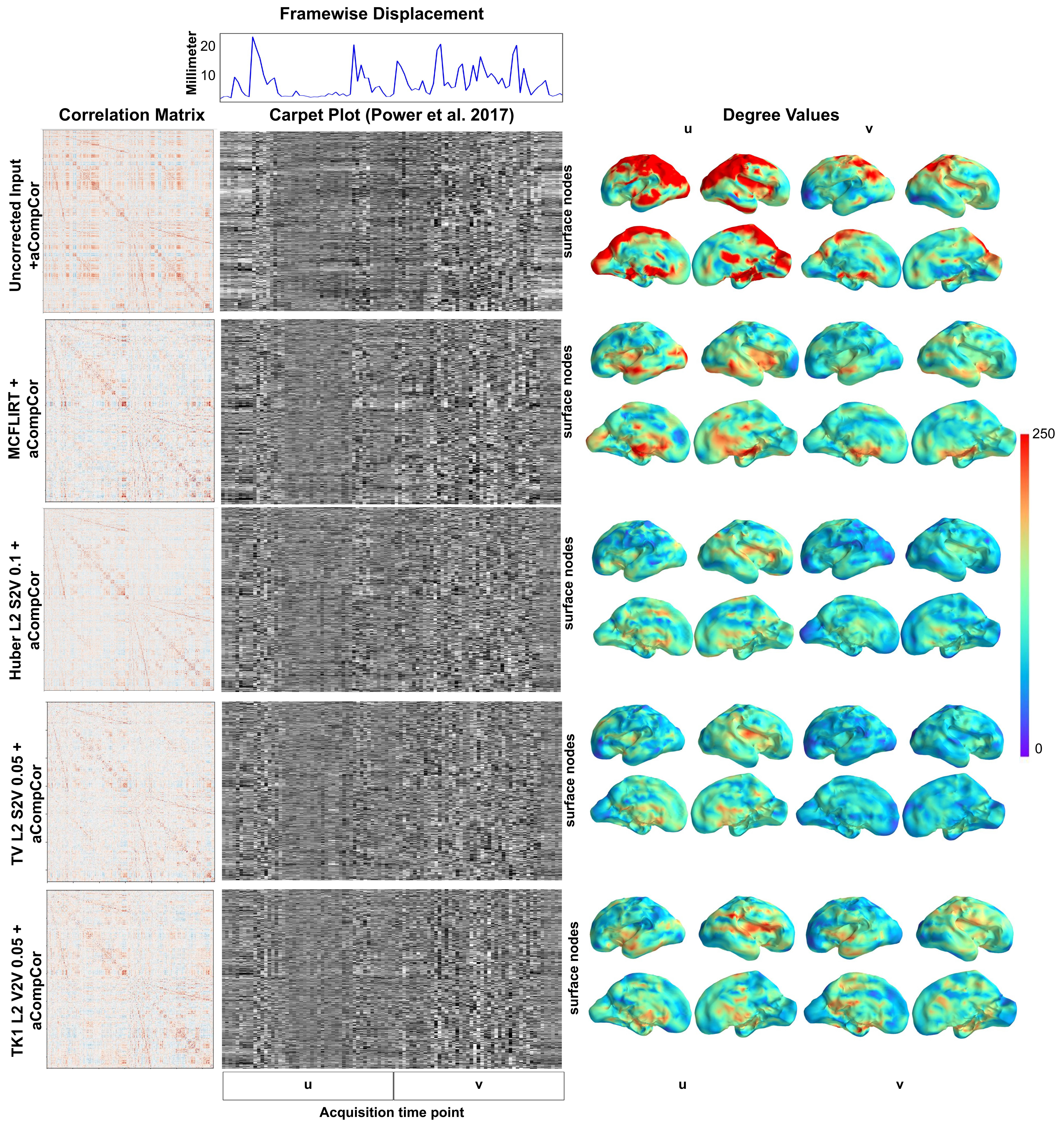}
\caption{Correlation matrix (column 1), carpet plots (column 2) and connectivity values (column 3) for a gestation week 25 subject with different processing steps} 
\label{fig:degree-conn}
\end{figure*}

Figure \ref{fig:boxplots} shows on the bottom right box the \textbf{correlation differences} for S2V and V2V motion correction with different regularization parameter values for TK1 L2, TV L2 and Huber L2 reconstruction, the uncorrected input and \\ MCFLIRT after aCompCor. The results of the correlation differences correlate with the results from the outlier ratio estimation (bottom left box in Figure \ref{fig:boxplots}) for Huber L2, where the best parameter values are $0.05$ and $0.1$. For further connectivity calculations and analysis we decided to show results for the best regularization parameter of each regularizer (Huber L2: S2V with $0.1$, TV L2: S2V with $0.05$, TK1 L2: V2V with $0.05$). \\
After motion correction and regularization the \textbf{QC-FC} correlations are improved, resulting in decreased distance dependency of FC values with subject motion (slope values: $-0.0029$ for uncorrected input, $-0.0010$ for Huber L2 $0.1$ S2V, $-0.0014$ for TV L2 $0.05$ S2V, $-0.0015$ for TK1 L2 $0.05$ V2V), where Huber L2 $0.1$ S2V performs best.\\
In Figure \ref{fig:degree-conn} \textbf{correlation matrices} (column 1), \textbf{carpet plots} (column 2) and \textbf{degree} value surface visualisations (column 3) for a subject at gestation week 25 (more examples are provided in the supplementary material in Figure S6 and S7) are demonstrated to show the effect of the proposed motion correction and volumetric reconstruction framework on signal interpretability with different regularization parameters. Carpet plots \cite{power2017simple} visualise for every surface node ($y$-axis) the intensity changes over time ($x$-axis) to assess the quality of an fMRI acquisition. 
 At the top of Figure \ref{fig:degree-conn} the framewise displacement \cite{power2012spurious} over time is plotted to show motion during the acquisition, where the $x$-axis (time) is shared with the carpet plot visualisation $x$-axis. It is observable that peaks of the framewise displacements correlate with edges in the carpet plot in row 1. Vertical (shifted intensities at one timepoint through motion) and horizontal edges (intensity changes over time) are observable in the uncorrected input without motion correction. These horizontal and vertical edges are reduced after motion correction (row 2 to row 5), since the fetal head is realigned and corresponding nodes are in the correct position as in the timepoints before and after. All three regularization methods (row 3 to row 5) show better results as the compared MCFLIRT motion correction (row 2). Overall, Huber L2 and TV L2 are very similar. We are favoring Huber L2 with smoother transitions and smoother edges (row 3). The correlation matrix shows higher proportions of motion (high correlation values across the whole matrix with \textit{checkerboard patterns}) in contrast to motion corrected (row 3 to row 5). Since we hypothesize that the two time ranges ($u$ and $v$) are more reproducible and similar after motion correction, consequently it is assumed that the degree value computation and corresponding visualisations for each of the time courses, illustrates similar patterns after successfully removing motion artefacts. Without motion correction (row 1) different patterns between $u$ and $v$ with high degree values are exposed, indicating that head movement introduces additional correlations. MCLFIRT (row 2) reduces this effect, but with Huber L2 S2V (row 3), TV L2 S2V (row 4) and TK1 L2 (row 5) these degree patterns look more similar and reproducible.
 \begin{figure*}%[!h]
\centering
\includegraphics[width=\textwidth]{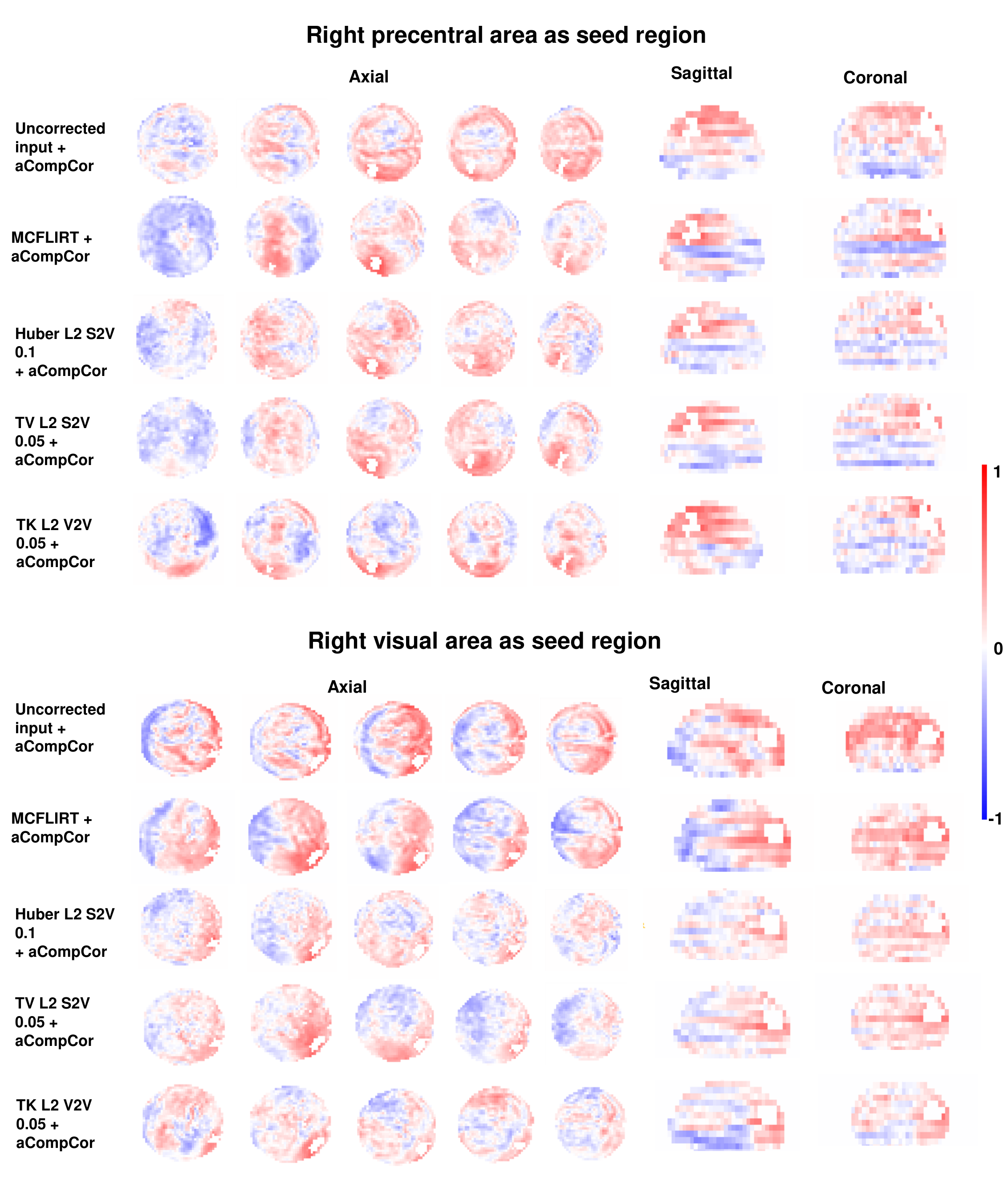}
\caption{\textcolor{Black}{Correlation values from right precentral area and right visual are as seed region (visualized as white area inside brain) versus other brain voxels. In the axial view we showed five consective slices and in the coronal and sagittal view a cross-section inside the seed region. Motion correction with Huber L2 and TV L2 regularization removed high correlation values caused by motion on border of brain mask and correlations are reduced and more limited to nearby seed regions.}} 
\label{fig:seed_corr}
\end{figure*}
\subsection{Experiment 4: Seed-based Functional Connectivity Analysis}
 \textcolor{Black}{As last experiment we performed seed-based functional connectivity analysis using seeds defined by the CRL\footnote{\url{http://crl.med.harvard.edu/research/fetal\_brain\_atlas/}} brain atlas. This atlas consists of 39 regions per hemisphere and was aligned with a subject's structural and functional MRI scan as illustrated in Figure \ref{fig:pp}. The aim of this analysis is to find regions, which activity pattern correlates with the activity pattern in a defined seed region. Therefore, the cross-correlation is computed between the activity sequence of the seed area and the activity time courses of the remaining brain voxels. As seed region we decided to evaluate the precentral area and occipital (visual) area, modules in fetal brain function, which have already been observed in former studies \cite{turk2019functional,jakab2014fetal,schopf2012watching}. In Figure \ref{fig:seed_corr} the seed-based analysis results are visualised for the same subject at gestation week 25, for which the surface-based analysis are illustrated in Figure \ref{fig:degree-conn}. On the upper half of Figure \ref{fig:seed_corr} the seed correlation values between the right precentral area and the rest of the brain are presented after performing aCompCor on the uncorrected input (row 1) and the reconstruction approaches using MC FLIRT (row 2), Huber L2 (row 3), TV L2 (row4) and TK1 L2 (row 5).}   
\textcolor{Black}{
 The lower half of Figure \ref{fig:seed_corr} shows the seed correlation values between the right visual area and the rest of the brain. Also here we evaluated 5 setups after performing aCompCor on the uncorrected input (row 1) and the reconstruction approaches using MC FLIRT (row 2), Huber L2 (row 3), TV L2 (row4) and TK1 L2 (row 5). }
 \subsubsection{Results}
  \textcolor{Black}{For both seed regions it is observable for uncorrected samples, that an increased number of high correlating regions are visible, including more voxel lying outside the expected functional area of precentral or visual area respectively and specifically at the border of the brain. This observation mirrors the expected erroneous signal correlations caused by motions \cite{esteban2019fmriprep}. This effect is reduced by all applied motion correction techniques, however in different ways: While for MCFLIRT correlations at the brain border are reduced, an increase of correlations in the border area of the seed region is observable as well larger localized areas of high correlations. In contrast to this Huber L2, TVL2 show only extrem high correlating regions in the regions nearby the seed region and do not radiate across the brain as observable for TKL2 and MC FLIRT.}   
\\ 

\section{Discussion}\label{sec:Discussion}
\textcolor{Black}{In this work we performed a profound parameter study of the proposed novel motion correction and reconstruction approach for fetal fMRI. \\
First, we aimed on assessing the motion correction performance (experiment 1) disentangled from the reconstruction process, to be able to test the approach's capability to correct for simulated defined translational or rotational movements.
We specifically focused on evaluating extreme parameter ranges in this experiment, spanning from movements smaller then the actual voxel resolution (<2.5 mm,<1°) to extremes of $\pm14$ degrees rotation or $\pm8$ mm of translation between to time points. Despite the choice of this challenging setup, we can report for 1.5T as well as 3T acquisitions stable motion correction performance for the introduced S2V technique with mean absolute angle errors between 0.2 and 0.4 degrees and mean absolute translation errors between 0.2 and 0.7 mm, outperforming V2V and MC FLIRT (see Figure \ref{fig:moco_eval}).}
\textcolor{Black}{We noticed that exact brain masking led to better motion estimation results compared to a dilated brain mask, which is used over the whole time series.}
The second experiment investigated the effect of regularization on the reconstructed image with different metrics. First, a parameter study (from $0.001$ to $1000$) was conducted to find the optimal parameters. Finally, the parameter value range from $0.005$ to $0.5$ was investigated for TK1 L2, from $0.05$ and $2$ for Huber L2 and from $0.01$ to $1$ for TV L2. 
\textcolor{Black}{For assessing the effect of regularization on the raw signal we developed and computed 4 benchmark metrics: (1) Standard deviation (2) SSIM (3) Outlier ratio and (4) Correlation difference.}
Results of the second experiment showed that standard deviation and SSIM metrics are sensitive to smoothing. In contrast, the outlier ratio metric was less sensitive to this effect and the corresponding results correlate with the correlation differences findings. Huber L2 and TV L2 regularization performed better as TK1 L2 regularization with a lower standard deviation, higher structural similarity and decreased outlier ratio (less rejected time points per subject). \textcolor{Black}{Therefore, it appeals to interpret the results from the standard deviation and SSIM metric only in combination with outlier ratio or connectivity differences.}\\
In the third experiment, the surface-based effect from motion correction and regularization was evaluated. Therefore, correlation differences are used as a metric to assess reproducibility of functional connectivity computations. The corresponding results demonstrated that Huber L2 regularization with $0.1$ and TV L2 regularization with $0.05$ was preferred for the reconstruction of fetal fMRI after motion correction, since decreased correlation differences between the two time ranges are noticeable in contrast to first-order Tikhonov. Further, it is demonstrated that the signals of two time ranges of a subject are more similar after the proposed motion correction and volumetric reconstruction framework, compared to the uncorrected input confirming our hypothesis that motion has been removed successfully by the method proposed. After motion correction and Huber L2 $0.1$ regularization based reconstruction, the outlier ratio metric decreased from $12.9\%$ to $8.6\%$ (p-value 0.019) of excluded time frames of a subject. Furthermore, it is demonstrated that the relation between subject movement and functional connectivity is decreased (slope value from -0.0029 to -0.0010) after motion correction. It has been shown that the reproducibility of functional connectivity estimates improved in the reconstruction procedure with aCompCor. \\
\textcolor{Black}{In the fourth experiment the volumetric effect from motion correction and regularization was evaluated by seed-based analysis. We selected the precentral and occipital (visual) area as seed regions, which have been already observed in former studies \cite{turk2019functional,jakab2014fetal,schopf2012watching}. Results showed that without motion correction an increased number of high correlating regions are visible especially at the border of the brain. With our approach and Huber L2 or TV L2 regularization high correlating regions are nearby the seed region and do not radiate across the brain as with TK1 L2 regularization or MCFLIRT. Further, correlations at the border due to motion artifacts are reduced.}
\\\\In summary Huber L2 regularization with a parameter of $0.1$ and S2V motion correction showed the best and most reproducible results on fetal fMRI data. 
\textcolor{Black}{The proposed framework is not limited to axial fMRI sequences and works also for coronal or sagittal recorded sequences.}
\textcolor{Black}{Locally non-rigid head deformations during the fMRI acquisition are not compensated by the approach proposed, being one of the current limits and the focus of future investigations. We observed that such deformations are a potential source of introduced artefacts in the reconstruction process and can influence the result of the reconstructed time series.}\\ 

\textcolor{Black}{We perceived that there is a higher amount of outlier time points in the younger aged fetal cases after our motion correction and volumetric reconstruction application, which can be explained by the fact that the fetuses in that age group have still more space to move in comparison to older fetuses. Additionally, the EPI and interleaved protocol in combination with extreme movements with framewise displacements larger 10 mm, showed a decreased reconstruction quality of our approach. However, the most displacements observed in our datasets were between $0$ and $10$ mm between two time points, while rare extreme movements reached up to 50 mm. Despite the challenging setup our proposed approach (using S2V motion correction and reconstruction with Huber L2 $0.1$ regularization) was capable to reduce the percentage of outlier timepoints in the data from $18.6\%$ to $9.16\%$ between GA $20$ and $25$, from $7.125\%$ to $5\%$ between GA $26$ and $31$ and from $9.33\%$ to $4.5\%$ between GA $32$ and $37$. 
} \\\\
In general with the proposed framework existing motion containing fMRI acquisitions are more usable since motion corrupted time points do not need to be excluded. Clinically, this has several advantages: Time consuming fetal fMRI data acquisitions can be held up to a necessary minimum to a more efficient use of existing data with our method. The image acquisition time can be kept to a certain limit, allowing for more compliance and comfort during the generally psychologically challenging diagnostic fetal MRI examinations of the pregnant women. More robust outlier rejection leads to less motion corrupted data and ultimately a more reliable single subject analysis, which is clinically highly desirable for the establishment of prognostic \textcolor{Black}{noninvasive imaging biomarkers \cite{thomason2017weak,thomason2019prenatal,de2020functional,van2021maternal}}.
\section{Conclusion}\label{sec:Conclusion}
In this work, a motion correction and volumetric reconstruction framework for 4D in-utero fMRI imaging data is presented, consisting of three components: (1) outlier-robust high-resolution reference volume construction, (2) a rigid slice-based motion correction and (3) a Huber L2-based volumetric reconstruction component. Synthetic data was created to simulate motion patterns to show that the proposed motion correction and volumetric reconstruction algorithm works with high motion corrupted slices. Furthermore, Huber L2 as regularization term for fetal fMRI motion correction was introduced and a parametric study for TK1 L2, TV L2 and Huber L2 regularization was conducted, to find the optimal reconstruction parameter value. A signal quality assessment strategy was presented, with intensity, outlier and connectivity based metrics to determine the impact of the proposed motion correction and volumetric reconstruction framework on functional connectivity computation. Further, we introduced a motion estimation evaluation scheme for synthesized data. Neither bias field correction nor other preprocessing steps are performed to avoid altered signals and noise properties in the comparison between the raw signal and the motion corrected signal. Our approach provides several advantages, especially in the setting of clinically diagnostic fetal fMRI. An optimal use of existing fMRI data, as proposed here, helps to avoid extensive and tedious BOLD sequence acquisitions. Further work is needed to evaluate its performance for clinical prognostic use of fetal resting state fMRI. \textcolor{Black}{We see our work as benchmark for the performance assessment of developed motion correction pipelines and intend to release an integrated analysis pipeline including our used quantification metrics in the future.}

\section*{Acknowledgement}
This work was supported by EU H2020 Marie \\Sklodowska-Curie [765148], Austrian Science Fund FWF [P 35189], Vienna Science and Technology Fund WWTF [LS20-065], The Wellcome Trust [WT101957; \\203148/Z/16/Z], the Engineering and Physical Sciences Research Council [NS/A000027/1; NS/A000049/1] and Novartis Pharmaceuticals Corporation.
\textcolor{Black}{\section*{Data and code availability statement}
The processed fMRI data was acquired under ethics number 790/2010 at the Medical University of Vienna and is not publicly available due to patient data protection. The code of the motion correction and volumetric reconstruction framework is available in release version 0.9 at \url{https://github.com/gift-surg/NiftyMIC}.}

\bibliographystyle{IEEEbib}
\bibliography{cas-refs}
\cleardoublepage
\onecolumn
\section*{Supplementary material}
\renewcommand{\thefigure}{S\arabic{figure}}
\setcounter{figure}{0}
\begin{figure}[!htb]
\centering
\includegraphics[width=1\textwidth]{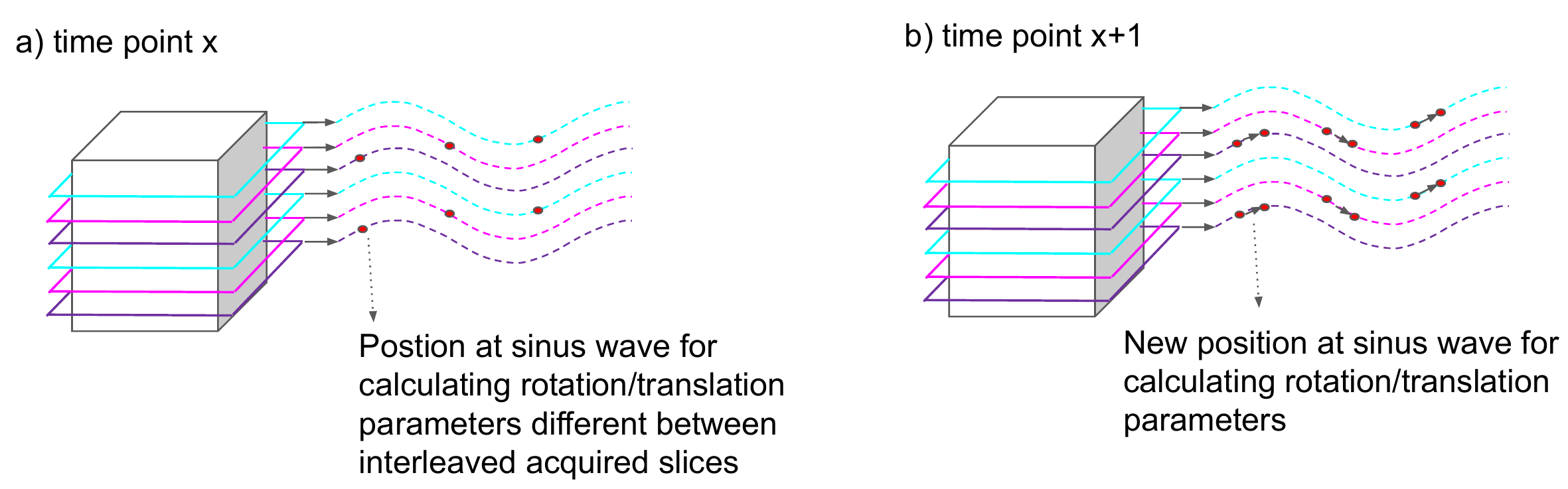}
\caption{\label{fig:s1}Imitation of interleaved movement patterns with a sinusoidal wave: a) Slices from one time point acquired at the same time have identical positions at the sinusoidal wave. b) At a subsequent time point the position on the sinusoidal wave changes.}
\end{figure}
\begin{figure}[!htb]
\centering
\includegraphics[width=1\textwidth]{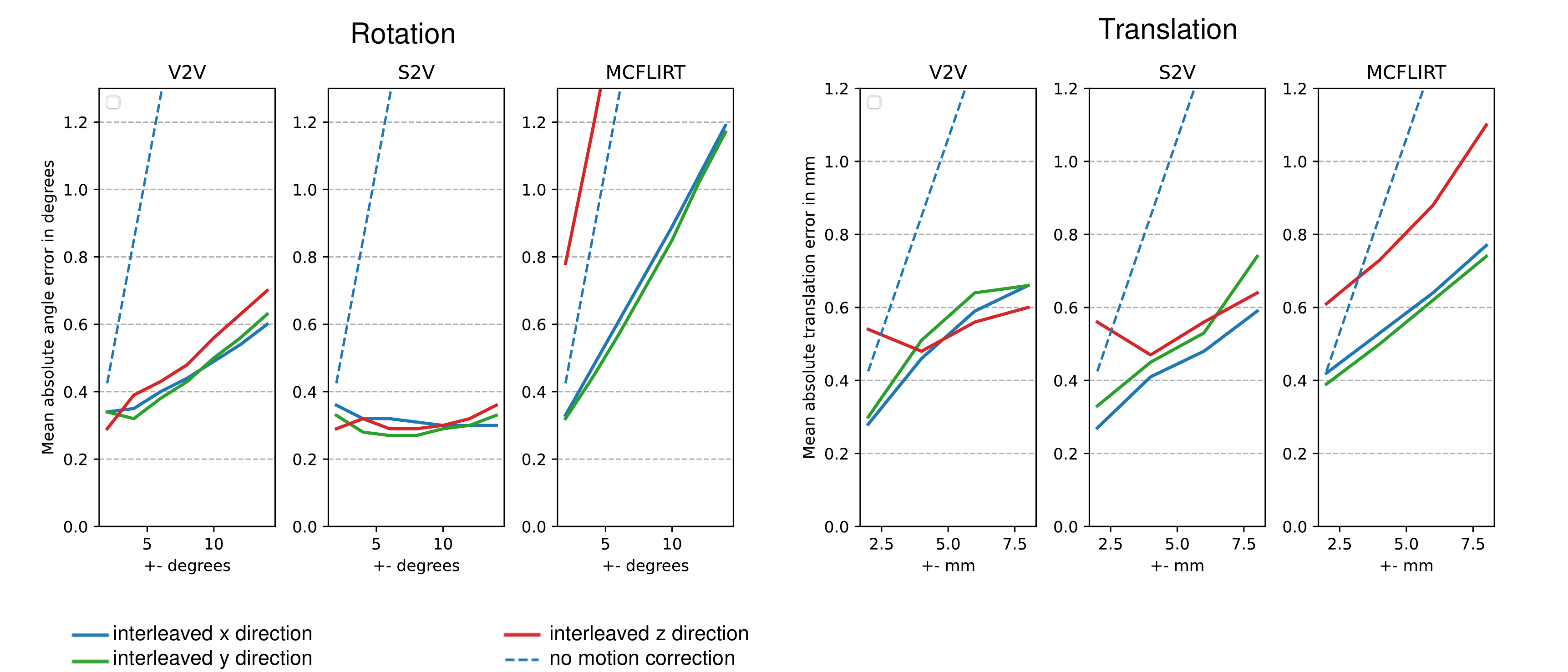}
\caption{\label{fig:s2}Visualization of the Mean Absolute Error (MAE) for obtained rotation (in degrees) and translation (in mm) parameters for V2V, S2V and MCFLIRT for the 3T synthetic data. The blue line shows the MAE in $x$ direction, the green line in $y$ direction and the red line in $z$ direction. The blue dashed line shows the MAE if no motion correction is performed.}
\end{figure}
\begin{figure*}
\centering
\includegraphics[width=1\textwidth]{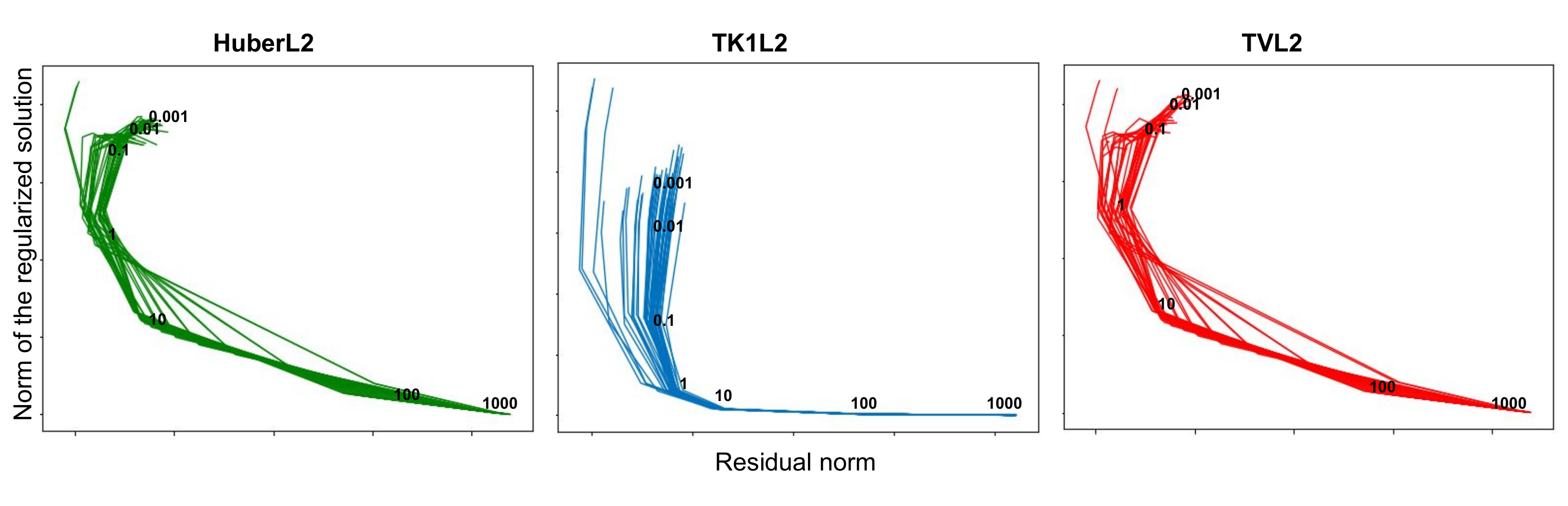}
\caption{\label{fig:s3}L-curve studies. Each curve shows one time point of a dataset. For one example curve the regularization parameters are visualised for Huber L2, TK1 L2 and TV L2.}
\end{figure*}
\begin{figure*}
\centering
\includegraphics[width=0.8\textwidth]{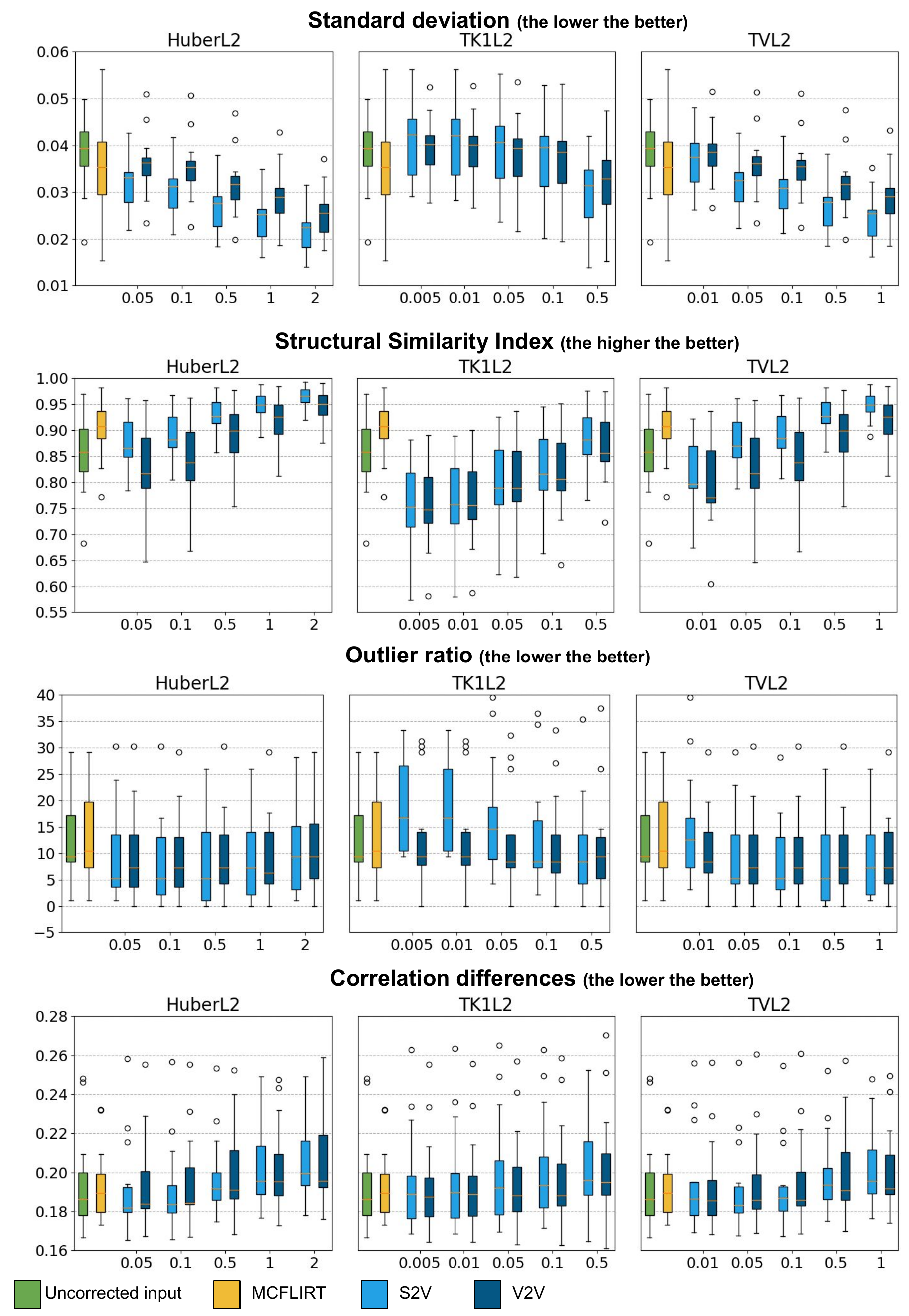}
\caption{\label{fig:s4}Boxplots for the four investigated metrics for 1.5 T acquisitions over the whole population for multiple motion correction strategies for different regularization parameters associated with TV L2, TK1 L2 and Huber L2.}
\end{figure*}

\begin{figure*}
\centering
\includegraphics[width=1\textwidth]{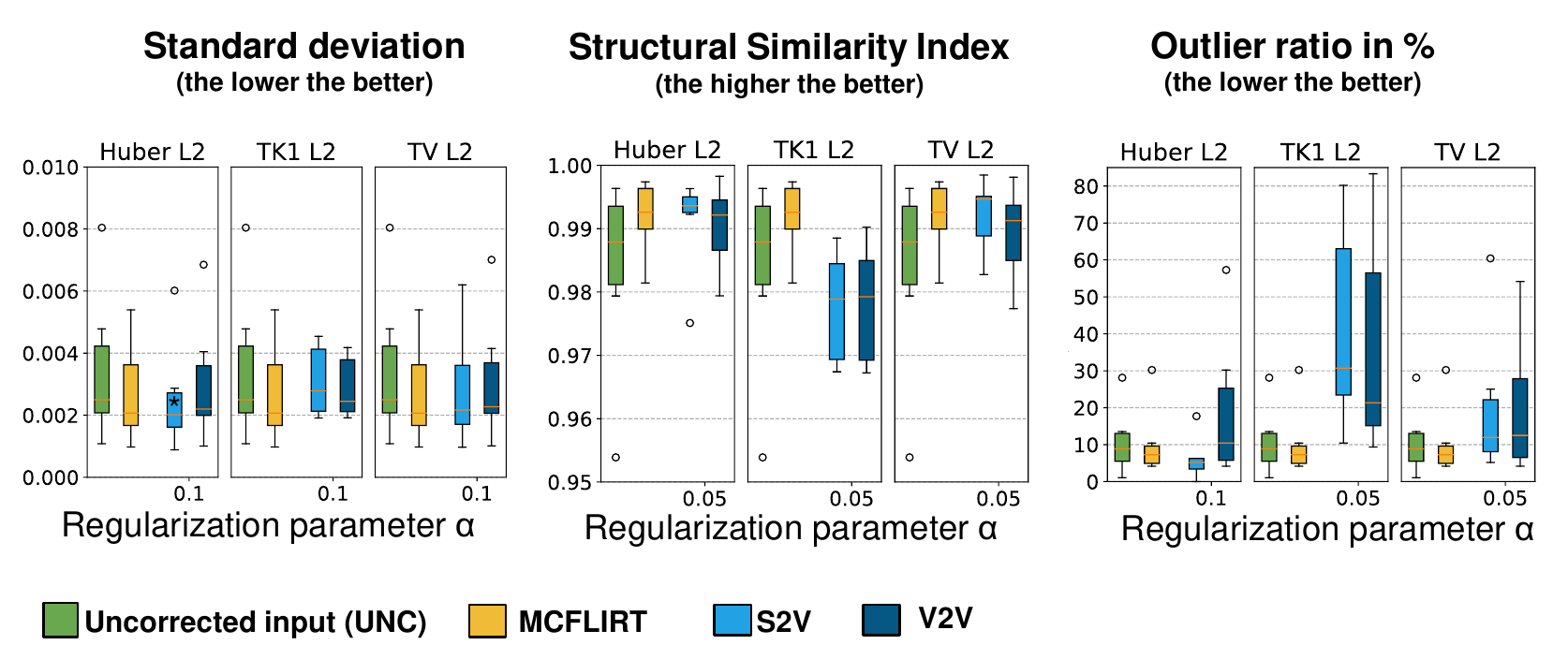}
\caption{\label{fig:s5}Results for standard deviation, structural similarity index and outlier ratio for the 3T subjects for multiple motion correction strategies (uncorrected input (UNC), MCFLIRT, S2V and V2V). We have chosen the best regularization parameter from experiment 2 associated with Huber L2, TK1 L2 and TV L2. Asterix symbols indicate statistical significance between UNC.}
\end{figure*}
\begin{figure*}
\centering
\includegraphics[width=1\textwidth]{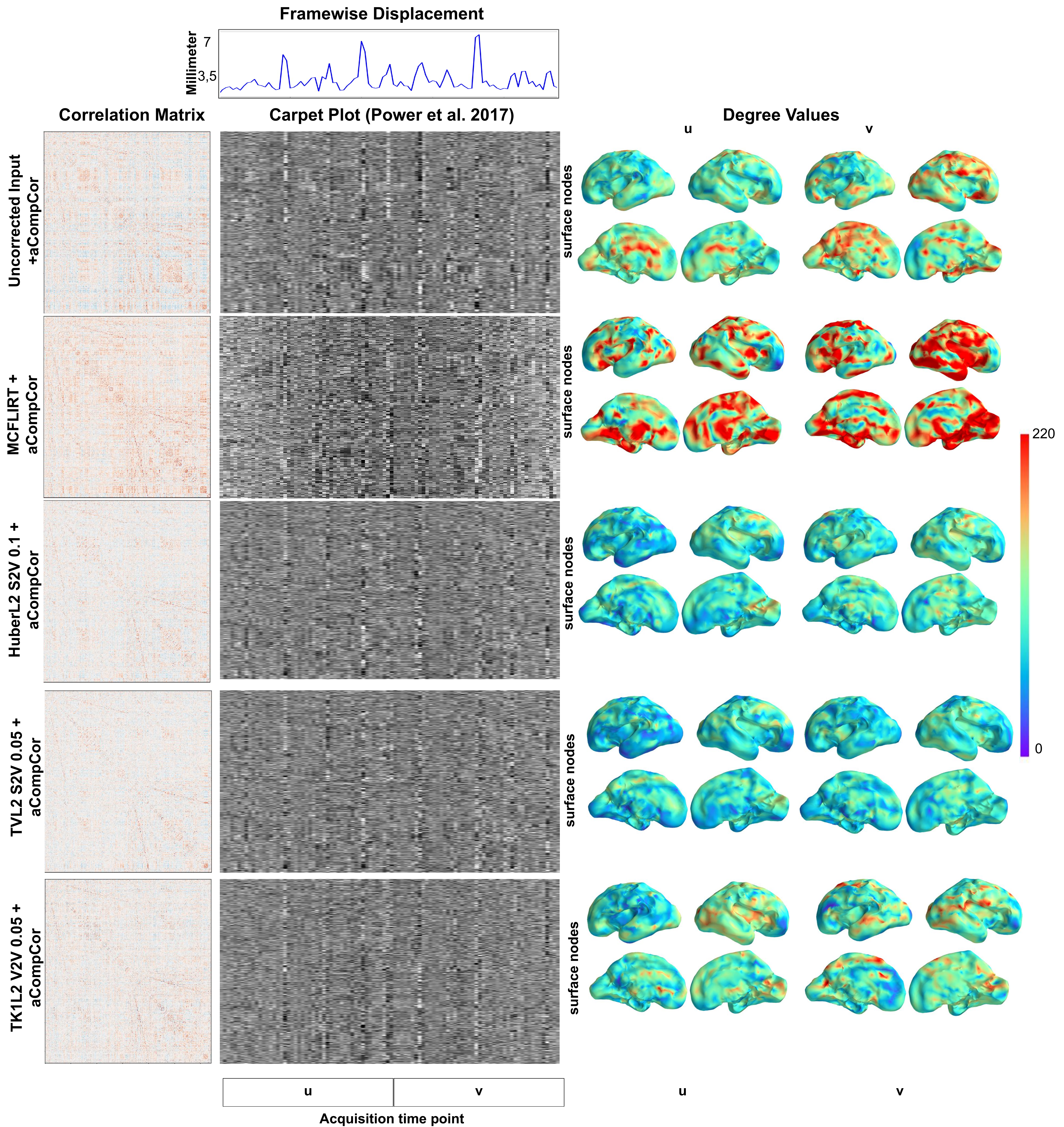}
\caption{\label{fig:s6}Correlation matrix (column 1), carpet plots (column 2) and connectivity values (column 3) for a gestation week 30 subject with different processing steps.}
\end{figure*}
\begin{figure*}
\centering
\includegraphics[width=1\textwidth]{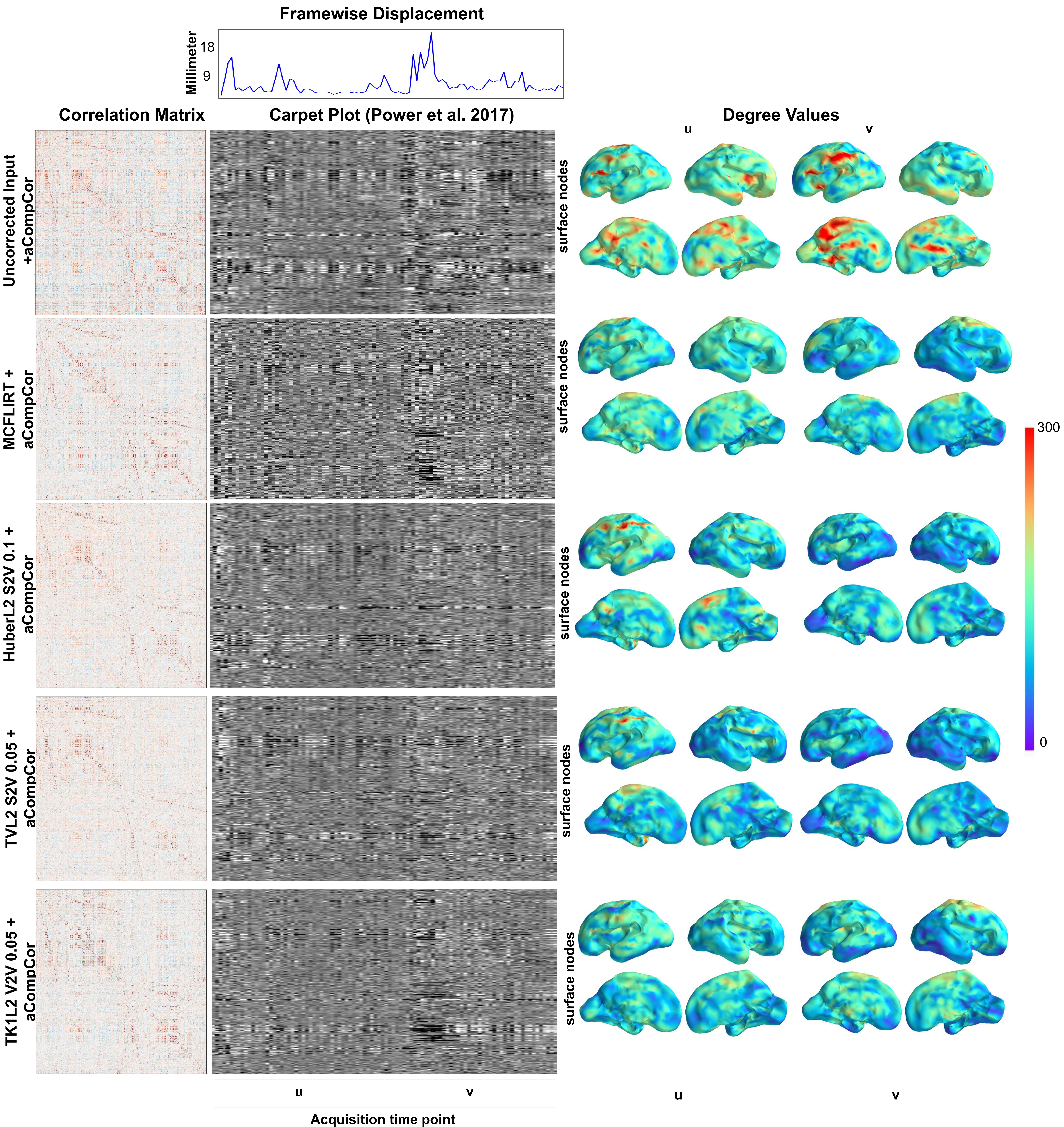}
\caption{\label{fig:s7}Correlation matrix (column 1), carpet plots (column 2) and connectivity values (column 3) for a gestation week 24 subject with different processing steps.}
\end{figure*}
\end{document}